\documentclass[10pt,twocolumn,letterpaper]{article}

\usepackage[pagenumbers]{cvpr} %

\usepackage[accsupp]{axessibility}

\usepackage{booktabs}
\usepackage{multirow}
\usepackage{tabularx}

\makeatletter
\renewcommand{\paragraph}{%
  \@startsection{paragraph}{4}%
  {\z@}{2.0ex \@plus 1ex \@minus .2ex}{-1em}%
  {\normalfont\normalsize\bfseries}%
}
\makeatother

\newcommand{\PAR}[1]{\vskip4pt \noindent {\bf #1~}}
\newcommand{\PARbegin}[1]{\noindent {\bf #1~}}

\usepackage{tikz}
\usetikzlibrary{positioning, calc, spy}
\usepackage{pgfplots}
\usepackage{pgfplotstable}
\pgfplotsset{compat=newest}
\usepgfplotslibrary{colorbrewer}
\usepgfplotslibrary{fillbetween}
\usepgfplotslibrary{groupplots}

\pgfplotstableset{
  y name/.estore in=\yname, 
  y err name/.estore in=\yerrname, 
}
\newcommand{\adderrorzone}[4][]{
\pgfplotstableset{#1}

\addplot[name path=upper2, draw=none, forget plot] table[y expr=(\thisrow{\yname} + \thisrow{\yerrname})] {#2};
\addplot[name path=lower2, draw=none, forget plot] table[y expr=(\thisrow{\yname} - \thisrow{\yerrname})] {#2};
\addplot[color=#4, semitransparent, forget plot, draw=none] fill between[of=upper2 and lower2];

\addplot[color=#3, thick] table[#1, y=\yname] {#2};
}

\tikzset{
    SVGGTMarker/.style={
        only marks, thick, blue, mark=x,
    },
    SPIMarker/.style={
        only marks, thick, red, mark=star,
    },
    SMAMarker/.style={
        only marks, very thick, orange, mark=Mercedes star,
    },
    VGGTMarker/.style={
        thick, blue, loosely dashdotted,
    },
    PIMarker/.style={
        thick, red, loosely dashdotted,
    },
    MAMarker/.style={
        thick, orange, loosely dashdotted,
    },
    FasterMarker/.style={
        thick, violet, dashed,
    },
    CuterMarker/.style={
        thick, green!90!black, dashed,
    },
    FlareMarker/.style={
        thick, yellow!80!black, dashed,
    },
}

\definecolor{cvprblue}{rgb}{0.21,0.49,0.74}
\usepackage[pagebackref,breaklinks,colorlinks,allcolors=cvprblue]{hyperref}

\def\paperID{41304} %
\def\confName{CVPR}
\def\confYear{2026}

\title{Block-Sparse Global Attention for Efficient Multi-View Geometry Transformers}

\author{%
Chung-Shien Brian Wang$^*$
\qquad
Christian Schmidt$^*$
\qquad
Jens Piekenbrinck
\qquad
Bastian Leibe\\[1em]
Computer Vision Group\\RWTH Aachen University\\
{\small \url{https://vision.rwth-aachen.de/sparse-vggt}}
}

\begin{document}
\maketitle
\begin{abstract}
Efficient and accurate feed-forward multi-view reconstruction has long been an important task in computer vision.
Recent transformer-based models like VGGT, $\pi^3$ and MapAnything have demonstrated remarkable performance with relatively simple architectures.
However, their scalability is fundamentally constrained by the quadratic complexity of global attention, which imposes a significant runtime bottleneck when processing large image sets.
In this work, we empirically analyze the global attention matrix of these models and observe that the probability mass concentrates on a small subset of patch-patch interactions corresponding to cross-view geometric correspondences.
Building on this insight and inspired by recent advances in large language models, we propose a training-free, block-sparse replacement for dense global attention, implemented with highly optimized kernels.
Our method accelerates inference by more than $3\times$ while maintaining comparable task performance.
Evaluations on a comprehensive suite of multi-view benchmarks demonstrate that our approach seamlessly integrates into existing global attention-based architectures such as VGGT, $\pi^3$, and MapAnything, while substantially improving scalability to large image collections.

\end{abstract}

\section{Introduction}
\label{sec:intro}
\renewcommand{\thefootnote}{\fnsymbol{footnote}}
\footnotetext[1]{Equal contribution.}
\renewcommand{\thefootnote}{\arabic{footnote}}

Reconstruction of 3D geometry and camera motion from multiple images is a central, long-standing problem in computer vision that has a broad impact across domains such as autonomous driving, embodied agents, AR/VR or photogrammetry.
Classical pipelines approach this task using optimization-based explicit geometric modeling and iterative refinement, most notably bundle adjustment, yielding reliable results at the cost of significant runtime overhead~\cite{schonberger2016colmap,pan2024glomap}.
More recently, feed-forward models~\cite{wang2024dust3r,tang2025mvdust3rplus,leroy2024mast3r} have narrowed the gap to structure-from-motion (SfM) systems such as COLMAP~\cite{schonberger2016colmap}.
A prevalent concept in such models, originating with DUSt3R~\cite{wang2024dust3r}, is to estimate 3D geometry and relative cameras from image-pairs and subsequently consolidate predictions across multiple views.

The recently proposed \emph{Visual Geometry Grounded Transformer (VGGT)}~\cite{wang2025vggt} is a particularly simple and effective multi-view geometry estimation model, achieving state-of-the-art performance on reconstruction, pointmap estimation and point tracking.
A key architectural design element of VGGT is the use of global attention blocks in the decoder, which enables holistic, scene-level reasoning in a single forward pass.
However, the complexity of global attention grows quadratically with the number of input images, thereby becoming the dominant computational contributor even at moderate sequence lengths, as illustrated in \cref{fig:quadratic-complexity}, which limits scalability to larger image collections.

\begin{figure}
  \centering
  \begin{tikzpicture}
\begin{axis}[
    width=0.95\linewidth,
    height=5cm,
    xlabel={Number of frames},
    ylabel={Time (s)},
    ymin=0,
    stack plots=y,
    area style,
    enlarge x limits=false,
    ymajorgrids,
    axis lines=left,
    x axis line style=-,
    y axis line style=-,
    legend pos=north west,
    reverse legend,
]

\pgfplotstableread[col sep=comma]{
N,total,global_block,global_attn
1,0.048201278686523436,0.01906790405511856,0.01414492791891098
2,0.059224830627441404,0.024641248106956483,0.018474304020404817
4,0.09941248321533203,0.04467372786998749,0.03442425572872162
8,0.20488716125488282,0.10342032051086426,0.08428278422355652
10,0.26514471435546877,0.1410401930809021,0.11818060779571533
20,0.618921142578125,0.3872313270568848,0.34467516613006594
50,2.36684130859375,1.812526351928711,1.7111921310424805
70,4.11276318359375,3.343490936279297,3.202699249267578
100,7.62996240234375,6.534946166992188,6.334964416503906
150,15.9220224609375,14.279359130859374,13.981034423828126
200,27.279505859375,25.110424560546875,24.712959045410155
300,59.47469921875,56.21087231445313,55.61919799804687
}\datatable

    \addplot[fill=Paired-G, draw=none] table[x=N, y=global_attn] {\datatable}\closedcycle;
    \addlegendentry{Global Attention}

    \addplot[fill=Paired-A, draw=none] table[x=N, y expr=\thisrow{total}-\thisrow{global_attn}] {\datatable}\closedcycle;
    \addlegendentry{Patch, FA, FFN}

\end{axis}
\end{tikzpicture}
  \caption{
    \textbf{Runtime of VGGT's forward pass.}
    FA denotes frame-wise attention.
    As the number of input frames increases, global attention dominates the computational cost
    (measured with FlashAttention2~\cite{dao2023flashattentionV2} on an H100 GPU at resolution $518^2$).
    We propose to adapt a block-sparse attention method that considerably reduces the cost of Global Attention while preserving result quality.
    }
  \label{fig:quadratic-complexity}
\end{figure}

By inspecting the global attention maps during a forward pass of the model, visualized in \cref{fig:sparse-global-attention-matrix}, we observe that each token only attends to a small subset of tokens, leading to sparse attention matrices. %
This observation motivates our systematic analysis of these sparsity patterns, suggesting that a substantial portion of the computational complexity of the global attention layers can potentially be avoided without degrading task performance.

Building on these findings and inspired by recent advances in large language models~\cite{gao2024seerattention}, we propose a novel adaptive block-sparse attention mechanism that approximates the full attention map and computes attention using block-sparse kernels whose cost scales with the number of active blocks rather than the full quadratic size, ultimately accelerating inference.
The additional module is lightweight and does not require any optimization, avoiding the need of extra annotations or backward passes through the original VGGT model.
With extensive experiments across multiple datasets, we empirically demonstrate that our method achieves task performance on par with the original VGGT model, while substantially accelerating end-to-end inference speed more than $3\times$.

To further demonstrate the versatility of our approach, we apply the proposed block-sparse attention mechanism to the recent $\pi^3$ variant of VGGT, designed for permutation-invariant geometry estimation~\cite{wang2025pi}, and to MapAnything~\cite{keetha2025mapanything}, which differs architecturally from VGGT yet similarly employs global cross-view attention.
We observe that both models exhibit similar accuracy–efficiency trade-offs, underscoring the generality of our method.

Overall, our contributions are threefold: (1) an analysis of sparsity-patterns in global attention layers of multi-view geometry estimation transformer models; (2) a novel adaptive block-sparse attention mechanism to accelerate inference speed without requiring optimization; and (3) comprehensive empirical evaluations demonstrating matched accuracy with significantly reduced compute.

\section{Related Work}

\subsection{Feed-Forward Multi-View Reconstruction}
Traditional multi-view reconstruction pipelines follow the \emph{Structure from Motion (SfM)} paradigm, jointly estimating scene geometry and camera poses~\cite{hartley2003sfm,schonberger2016colmap,schoenberger2016colmap-mvs,pan2024glomap}.
Typical implementations perform feature extraction and matching, triangulation, and bundle adjustment.
An additional dense depth estimation stage is necessary if a dense reconstruction is desired.
COLMAP~\cite{schonberger2016colmap} remains a widely adapted pipeline and is frequently employed as baseline or used to produce ground-truth annotations for datasets.

Learning-based approaches for multi-view reconstruction aim to enhance or replace these pipelines.
A prominent line of work employs a cross-attention decoder to match two input frames~\cite{wang2024dust3r,leroy2024mast3r,duisterhof2024mast3r-sfm,jang2025pow3r}.
Processing multiple views requires additional global alignment post-processing that consolidates pairwise predictions into a single, consistent scene representation~\cite{wang2024dust3r,leroy2024mast3r,duisterhof2024mast3r-sfm}.
This consolidation is computationally expensive and introduces a separate source of error.

To mitigate the cost of globally aligning image-pairs, several derivative works incrementally extend the reconstruction~\cite{elflein2025light3r,wang2025cut3r,wang2024spann3r} by iteratively applying two-view models.
Some maintain a persistent latent representation to accumulate information~\cite{wang2025cut3r,wang2024spann3r}, while Light3r-SfM~\cite{elflein2025light3r} directly updates a global reconstruction and shows improved performance on large scenes.
Despite their practicality, iterative schemes may not perform as well as methods that consider all views simultaneously.
Incremental updates are order-sensitive, prone to error accumulation and drift, and 
require revisiting past states or additional optimization to enforce long-range, cross-view consistency.
In contrast, globally conditioned models can exploit holistic context to align distant views and resolve ambiguities in a single forward pass, albeit at higher computational cost.

\subsection{Global Attention for Fast 3D Processing}
In contrast to two-view models, architectures with global attention mechanisms are able to reason jointly over all views at once, enabling scene-level consistency~\cite{yang2025fast3r,cabon2025must3r,wang2025vggt,wang2025pi}.
Fast3r~\cite{yang2025fast3r} combines a per-frame feature encoder with a decoder composed exclusively of global attention blocks, hence lacking frame-wise attention blocks.
In addition, it predicts both local and global pointmaps for every view, a design inherited from two-view models.

\begin{figure}
  \centering
  \includegraphics[width=0.48\textwidth]{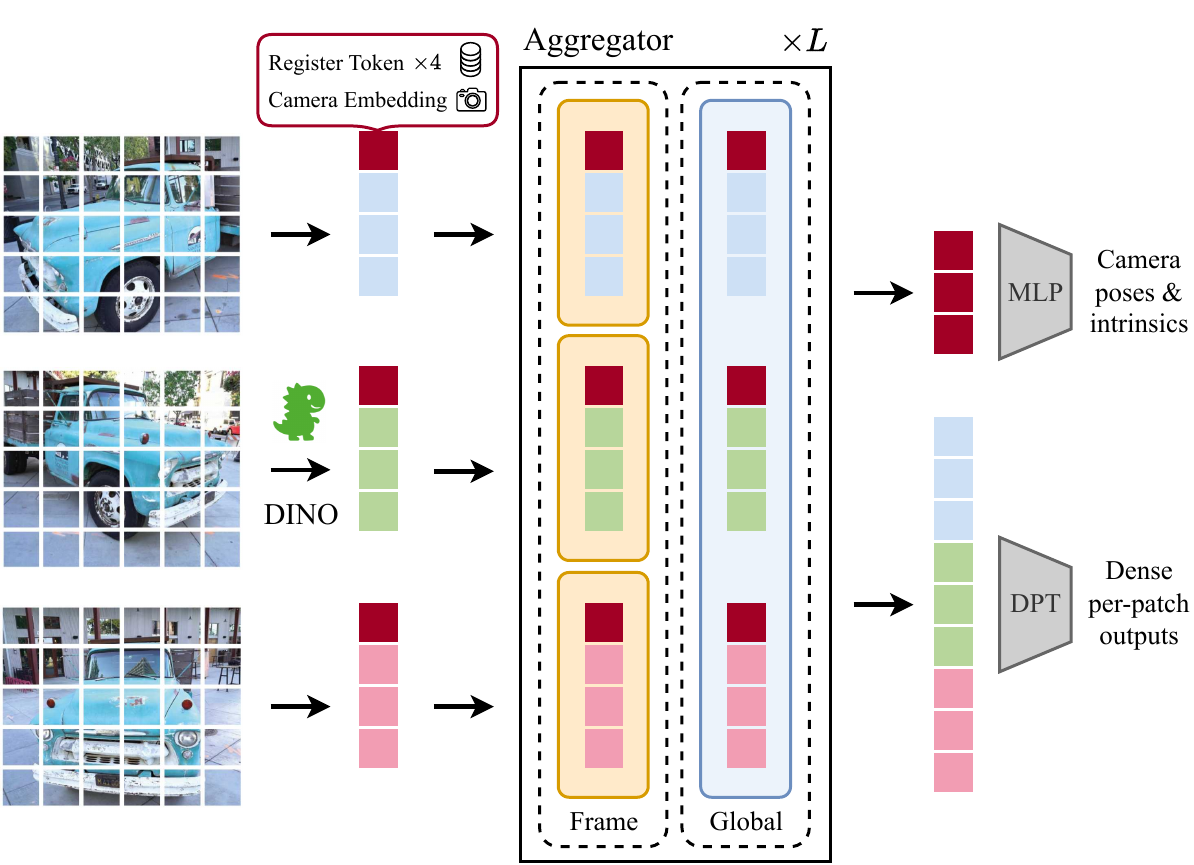}
  \caption{
    \textbf{Architecture overview of VGGT~\cite{wang2025vggt}.}
    The key component is the Aggregator consisting of $L = 24$ alternating attention blocks (first frame-wise attention, then global attention over all frames).
    Each input frame is augmented with five learned embedding vectors: one camera token and four register tokens.
    After the Aggregator, VGGT regresses camera poses from the camera tokens using a light-weight MLP head, and dense outputs (point maps, depth, point tracks) using DPT heads~\cite{ranftl2021dpt}.
    }
  \label{fig:vggt-architecture}
\end{figure}

VGGT~\cite{wang2025vggt} simplifies reconstruction by predicting all pointmaps in the coordinate system of a reference view, \ie the first view.
Its decoder alternates between frame-wise and global attention, allowing per-frame features to be adapted before the next round of global matching.
For reference, we provide an overview of the VGGT architecture in~\cref{fig:vggt-architecture}.
Since VGGT predicts all pointmaps with respect to the reference view and employs camera embeddings, its predictions are sensitive to input ordering.
To alleviate this issue, $\pi^3$~\cite{wang2025pi} removes the camera embeddings to attain permutation invariance while retaining per-frame register tokens akin to VGGT.
The authors report slightly improved reconstruction quality over VGGT on most benchmarks, with notably reduced variance across input permutations.
MapAnything~\cite{keetha2025mapanything} adopts a distinct transformer design for metric 3D reconstruction capable of accommodating optional geometric inputs such as intrinsics, poses, or partial depth alongside RGB images.

Despite their architectural differences, these methods share a reliance on global attention, whose time complexity scales quadratically with the number of tokens, which motivates our approach to exploit the sparsity of global attention to preserve reconstruction performance while reducing computational overhead.

\subsection{Sparse Attention}

Attention is the foundation of many state-of-the-art neural architectures, particularly large language models (LLMs).
Recent research has focused on reducing the quadratic complexity of dense attention by exploiting structure within the attention maps~\cite{chen2021scatterbrain,chen2022pixelated-butterfly,gao2024seerattention,zhang2025spargeattention}.
LLM-oriented sparse attention methods~\cite{jiang2024minference,lai2025flexprefill,yuan2025native,lu2025moba} rely on autoregressive assumptions, like causal masking or key-value caching during inference, that do not directly transfer to attention over continuous, two-dimensional token grids used in vision models.
Other works~\cite{zhang2025vsa,hassani2023neighborhood} leverage temporal or spatial continuity, assumptions that do not hold for global attention in multi-view vision transformers.

In contrast to weight sparsity and pruning approaches~\cite{frantar2023sparsegpt,sun2023wanda,yang2025wanda++} that compress model weights, our method targets attention sparsity, \ie sparsity emerging in the attention maps themselves.
Combining these two forms of sparsity could potentially lead to further acceleration.

Chen~\etal~\cite{chen2021scatterbrain} approximate attention by combining sparse and low-rank representations, but their irregular memory access pattern is inefficient on block-oriented accelerators.
A more generally applicable approach is block-sparse attention~\cite{chen2022pixelated-butterfly}, which combines the hardware-friendly, block-wise computation of FlashAttention~\cite{dao2022flashattention} with a sparse block mask.
While PixelatedButterfly~\cite{chen2022pixelated-butterfly} supports both weight and attention sparsity, it relies on static patterns and learned parameters that require optimization.
SeerAttention~\cite{gao2024seerattention} introduces a learned gating mechanism, which requires optimization, with a top-$k$ block selection strategy.
SpargeAttention~\cite{zhang2025spargeattention} recently showcased the effectiveness of block-sparse attention across diverse models and modalities.
It is training-free, combining self-similarity-based and CDF-thresholded block selection with a sparse online softmax computation.
To increase self-similarity and enable even sparser block selections, it applies Hilbert-curve permutations to spatially reorder tokens prior to attention.

Building on the block-sparse paradigm, we leverage the attention kernel of SpargeAttention~\cite{zhang2025spargeattention} to accelerate inference of VGGT~\cite{wang2025vggt}, $\pi^3$~\cite{wang2025pi} and MapAnything~\cite{keetha2025mapanything}. %
Unlike SpargeAttention, we omit self-similarity-based selection, Hilbert-curve permutation and the sparse online softmax.
Instead, we combine CDF-based and top-$k$ thresholding for block selection and retain dense attention for special tokens.
While Zhang~\etal~\cite{zhang2025spargeattention} reported results with sequences up to 128K tokens and $54\%$ sparsity under causal attention, our largest configuration processes sequences exceeding 512K tokens at a sparsity ratio above $75\%$.
We achieve speed-ups of more than $3\times$ while preserving task performances across all models.

\section{Analyzing VGGT's Attention Patterns}

Traditional multi-view reconstruction pipelines rely heavily on hand-crafted features, \eg SIFT~\cite{lowe2004sift}, in order to establish sparse correspondences that drive structure and camera estimation~\cite{schonberger2016colmap,schoenberger2016colmap-mvs}.
Our analysis of the global attention in VGGT reveals a strong concentration on a limited set of token pairs, resulting in sparse patterns that are analogous to correspondence matrices and that mirror geometric matching.
The key idea of our approach is to exploit these sparsity patterns to improve the efficiency of global attention.

\subsection{Background: VGGT's Architecture}
An overview of the VGGT architecture is illustrated in~\cref{fig:vggt-architecture}.
We briefly recapitulate the most important parts of the architecture for completeness.

VGGT consists of a patchifier $P$, a feature aggregator $A$, and several task-specific heads $H_1,\ldots,H_T$.
The patchifier $P$ independently transforms each input image $I$ into a set of patch tokens $T = P(I)$.
In VGGT, $P$ is implemented as a pre-trained DINOv2-Large~\cite{oquab2023dinov2,darcet2023register} with a patch size of $14\times 14$ pixels~\cite{wang2025vggt}.
For every frame, VGGT concatenates five special tokens to the patch tokens: a camera token and four register tokens~\cite{darcet2023register}.
To distinguish between the reference view and auxiliary frames, VGGT maintains one set of special tokens for the first frame and another set for all other frames.
The aggregator $A$ is a series of transformer blocks alternating between global and frame-wise attention.
Global attention computes full self-attention over the union of all patch and special tokens across all frames, which enables scene-level reasoning.
Frame-wise attention performs self-attention between the patch and special tokens of a single frame, which supports per-frame adaptation.
Both frame-wise and global attention layers are followed by two-layer per-token MLPs, as is common in transformer architectures~\cite{dosovitskiy2020vit}.
After the final aggregator layer, the register tokens are discarded and the remaining tokens are passed to the task-specific heads.
VGGT employs a lightweight head for pose regression and DPT~\cite{ranftl2021dpt} heads for dense prediction tasks, \ie depth and pointmaps.
Our analysis focuses on the global attention layers inside the aggregator. %

The follow-up model $\pi^3$~\cite{wang2025pi} closely matches VGGT's architecture with just a few modifications to promote permutation invariance.
Architecturally, $\pi^3$ uses fewer alternating attention blocks in the aggregator, $18$ instead of $24$, replaces DPT with plain Transformer heads for dense prediction tasks and removes camera embeddings while retaining per-frame register tokens.
Methodologically, $\pi^3$ supervises relative poses between all frames rather than coordinates tied to the reference view.
The authors show that these modifications improve the model's task performance and reduce sensitivity to input permutations~\cite{wang2025pi}.

\begin{figure*}
\centering
\centering

\begin{tikzpicture}[
    image/.style={
        inner sep=0pt,
        outer sep=1pt,
    },
    label/.style={
        anchor=south west, 
        outer sep=2pt,
        inner sep=2pt,
        color=black, 
        fill=white,
        opacity=0.8,
        text opacity=1,
        rounded corners=3pt,
        text height=1.5ex,
        text depth=0ex,
    },
    node distance=1pt,
    every node/.style={inner sep=0},
    spy using outlines={
        magnification=3,
        size=2.9cm,
        connect spies,
        every spy on node/.append style={ultra thick}
    },
]

\node[image] (Ltop) {\includegraphics[height=3cm]{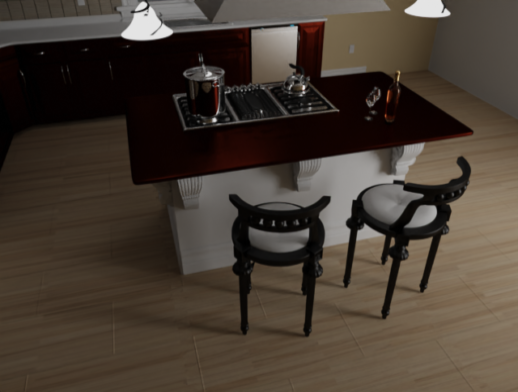}};
\node[label] at (Ltop.south west) {Frame 0};
\node[image, below=of Ltop] (Lbot) {\includegraphics[height=3cm]{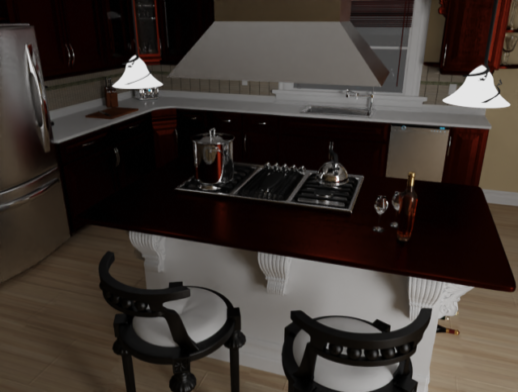}};
\node[label] at (Lbot.south west) {Frame 1};

\node[image, right=2mm of Ltop.north east, anchor=north west] (R) {\includegraphics[height=6.1cm]{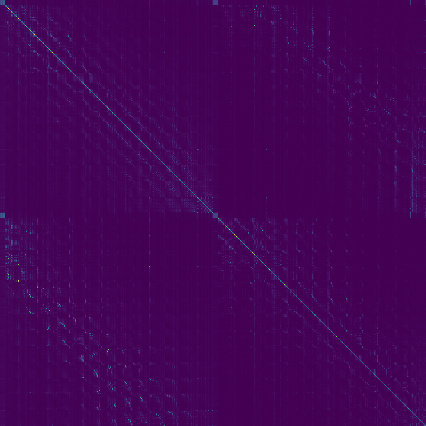}};

\spy[name=mag] on ($ (R.north east)!0.5!(R.south west) $) in node [image, right=2mm of R.north east, anchor=north west];
\spy on ($ (R.north east)!0.8!(R.south west) $) in node [right=2mm of R.south east, anchor=south west];

\node[right=3.4cm of R.north east, anchor=north west] {
  \begin{tikzpicture}
    \begin{axis}[
      hide axis, scale only axis, height=0pt, width=0pt,
      colorbar, colormap/viridis,
      point meta min=0, point meta max=0.43,
      colorbar style={
        height=6cm,
        yticklabel style={xshift=2pt},
        ylabel={Attention Score},
        y label style={anchor=north, yshift=-8pt}
    }
    ]
      \addplot [draw=none] coordinates {(0,0)};
    \end{axis}
  \end{tikzpicture}
};
\end{tikzpicture}
\caption{
    \textbf{Visualization of VGGT's global attention matrix.}
    A very small number of entries are highly activated, while the vast majority of entries are near zero.
    This visualization shows the average attention map over all heads of layer 15 in the VGGT aggregator, at an input resolution of $224\times 182$.
    Upper highlight: The special tokens attend to each other and form a distinctive pattern.
    Lower highlight: Patch-level attention is localized on a small subset of highly activated entries.
    See the supplementary material for an enlarged visualization.
}
\label{fig:sparse-global-attention-matrix}
\end{figure*}

\subsection{Visualizing Attention Maps}
\begin{figure*}
    \begin{minipage}{0.5\textwidth}
        \includegraphics[width=0.9\linewidth]{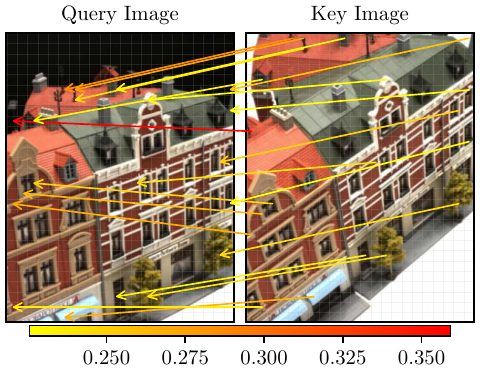}%
        \hfill
    \end{minipage}%
    \begin{minipage}{0.5\textwidth}
        \hfill%
        \begin{tikzpicture}
\pgfplotstableread[col sep=comma]{fig/attn_stats/attn_stats_vggt.csv}\tableAttnStats

\begin{groupplot}[
    group style={group size=1 by 2, vertical sep=4pt},
    width=8.3cm, height=3.7cm,
    ymode=log,
    xmin=0.1, xmax=24.9,
    xlabel={Global Attention Layer},
    ylabel={Activation},
    legend style={
        draw=none,
        font=\small,
        at={(0.5,-0.55)},
        anchor=north,
        legend columns=4
    }
]

\nextgroupplot[xticklabels=\empty, xlabel={}, ylabel={Max Entry}]

\adderrorzone[x expr=\coordindex+1, y name=s2sMaxMean, y err name=s2sMaxStd] {\tableAttnStats}{Paired-B}{Paired-A};
\adderrorzone[x expr=\coordindex+1, y name=s2pMaxMean, y err name=s2pMaxStd] {\tableAttnStats}{Paired-D}{Paired-C};
\adderrorzone[x expr=\coordindex+1, y name=p2pMaxMean, y err name=p2pMaxStd] {\tableAttnStats}{Paired-F}{Paired-E};
\adderrorzone[x expr=\coordindex+1, y name=p2sMaxMean, y err name=p2sMaxStd] {\tableAttnStats}{Paired-H}{Paired-G};

\nextgroupplot[ylabel={Avg. Entry}]

\adderrorzone[x expr=\coordindex+1, y name=s2sMeanMean, y err name=s2sMeanStd] {\tableAttnStats}{Paired-B}{Paired-A};
\addlegendentry{S2S}
\adderrorzone[x expr=\coordindex+1, y name=s2pMeanMean, y err name=s2pMeanStd] {\tableAttnStats}{Paired-D}{Paired-C};
\addlegendentry{S2P}
\adderrorzone[x expr=\coordindex+1, y name=p2pMeanMean, y err name=p2pMeanStd] {\tableAttnStats}{Paired-F}{Paired-E};
\addlegendentry{P2P}
\adderrorzone[x expr=\coordindex+1, y name=p2sMeanMean, y err name=p2sMeanStd] {\tableAttnStats}{Paired-H}{Paired-G};
\addlegendentry{P2S}

\end{groupplot}
\end{tikzpicture}
    \end{minipage}
    \caption{
    \textbf{VGGT's global attention matrix is extremely sparse.}
    Left: We visualize the tokens corresponding to the top-k activated entries of the attention map of layer 15.
    Right: Average \& maximum attention scores in the global attention maps; the shorthand \{S,P\}2\{P,S\} denotes attention between special (S) and patch (P) tokens.
    Solid lines denote the mean over all tokens, while shaded regions denote the standard deviation.
    Increasing indices indicate layers closer to the output.
    Layers in the middle of the aggregator exhibit higher activations and increased sparsity.
    Note the different scalings of the mean and max activations.
    }
    \label{fig:sparse-global-attention}
\end{figure*}

We visualize the full post-softmax attention map for a global attention block in a middle decoder layer, layer $15$, in~\cref{fig:sparse-global-attention-matrix}.
Only a small fraction of entries carry non-negligible probability mass, indicating pronounced sparsity.
To study how this behavior evolves with layer depth, \cref{fig:sparse-global-attention}~(right) plots the average and the maximum activation of different parts of the attention map against the layer index.
We observe a peak in the maximum activation values in the middle of the aggregator.
This effect is concentrated in patch-patch interactions, whereas attention involving special tokens remains stable across layers.
The sparse, highly activated entries align with geometrically meaningful correspondences between frames, as illustrated in~\cref{fig:sparse-global-attention}~(left).

We hypothesize that the model learns to perform an exhaustive correspondence search over the layers of the aggregator.
More concretely, the global attention proposes cross-view correspondences and exchanges information across frames, while the frame-wise attention adapts patch token features to improve matches in the subsequent global layers.
\cref{fig:sparse-global-attention}~(right) shows that the mid-aggregator layers exhibit the strongest selective attention and appear to contribute disproportionately to multi-view reasoning.

\begin{figure}
  \centering
  \begin{tikzpicture}

\pgfplotstableread[col sep=comma]{fig/layer_drop/layer_drop_both_co3d.csv}\tableSkipBoth

\pgfplotstableread[col sep=comma]{fig/layer_drop/layer_drop_front_co3d.csv}\tableSkipFront

\pgfplotstableread[col sep=comma]{fig/layer_drop/layer_drop_middle_co3d.csv}\tableSkipMiddle

\pgfplotstableread[col sep=comma]{fig/layer_drop/layer_drop_back_co3d.csv}\tableSkipBack

\begin{axis}[
    width=8.3cm,
    height=4cm,
    xmin=0, xmax=24.5,
    ymin=0, ymax=1.0,
    axis lines=left,
    xlabel={\#Skipped global attention},
    ylabel={AUC@30},
    tick label style={font=\small},
    label style={font=\small},
    legend style={
        draw=none,
        font=\small,
        at={(0.45,-0.4)},
        anchor=north,
        legend columns=-1,
        row sep=1pt,
    },
    legend cell align=left,
    clip=false,
]

\addplot[
  dashed,
  black,
  domain=0:24
] {0.9267};
\addlegendentry{None}

\addplot[mark=*, line width=1.1pt, color=Paired-B] table[x=numskipped, y=AUC]{\tableSkipFront};
\addlegendentry{Front}

\addplot[mark=*, line width=1.1pt, color=Paired-D] table[x=numskipped,y=AUC]{\tableSkipBack};
\addlegendentry{Back}

\addplot[mark=*, line width=1.1pt, color=Paired-F] table[x=numskipped,y=AUC]{\tableSkipBoth};
\addlegendentry{Front \& Back}

\addplot[mark=*, line width=1.1pt, color=Paired-H] table[x=numskipped,y=AUC]{\tableSkipMiddle};
\addlegendentry{Mid}

\end{axis}
\end{tikzpicture}
  \caption{
      \textbf{Influence of dropping global attention layers.}
      We skip the computation of different global attention layers in the aggregator starting with the earliest (Front), last (Back), alternating (Front \& Back), or from the middle layers (Middle), and evaluate pose estimation on CO3Dv2~\cite{reizenstein21co3d}.
      The x-axis denotes the total number of skipped layers.
      The experiment shows that the model is especially sensitive to pruning of the center layers, and robust against pruning the early and late layers.
  }
  \label{fig:layer-drop}
\end{figure}

\subsection{Are Some Layers More Important?}
To further test our hypothesis that the mid-layers of the aggregator are most important for task performance, we conduct a layer-drop ablation in which selected aggregator blocks are skipped at inference.
We show the ablation results in~\cref{fig:layer-drop} and show additional results in the supplementary material.
The figure shows that the model is comparatively robust when early or late aggregator layers are skipped.
In contrast, omitting a single middle layer leads to a pronounced performance drop.
This behaviour supports our hypothesis that the middle of the aggregator concentrates the essential cross-view information exchange and motivates our adaptive block-selection strategy.

\subsection{Consequences of this Analysis}

Our analysis yields three take-away messages.
First, global attention is highly sparse.
Second, the variation in attention entries is dominated by patch–patch interactions, while special token interactions remain stable across depth.
Lastly, mid-stack layers carry the key cross-view integration, as confirmed by the layer-drop ablation.
These findings indicate that dense quadratic attention is not necessary throughout the stack, but that considerable efficiency gains can be realized by exploiting sparsity, in particular in the important middle layers.
Guided by this analysis, we propose to replace the dense global attention layers with an adaptive block-sparse variant.
This reduces the time complexity with minimal accuracy loss and integrates into VGGT and $\pi^3$ without changes to their encoders or task heads.

\section{Method}
In this section, we describe our method for block-sparse attention in the global attention layer in VGGT (\cref{ssec:pooling}) and how we handle camera and register tokens (\cref{ssec:special-token-handling}).
Note that while we utilize the implementation of SpargeAttention~\cite{zhang2025spargeattention}, block-sparse attention in itself is independent of that particular work and could, \eg, be implemented on top of FlashAttention~\cite{dao2022flashattention,dao2023flashattentionV2,zhang2025flare} to yield similar results.
We provide a theoretical analysis of the achievable speed-up for a certain architecture in the appendix.

\subsection{Block-Sparse Attention}
\label{ssec:pooling}

\PARbegin{Self-Attention.}
Self-Attention is the cornerstone of many modern neural networks, as it enables dynamic, global interactions between input elements.
Given input tokens $\mathbf{X} \in \mathbb{R}^{n \times d}$, queries $\mathbf{Q}$, keys $\mathbf{K}$, and values $\mathbf{V}$ are computed as linear projections of $\mathbf{X}$, such that $\mathbf{Q}=\mathbf{XW}^Q$, $\mathbf{K}=\mathbf{XW}^K$, $\mathbf{V}=\mathbf{XW}^V$, and the scaled dot-product attention is defined as
\begin{equation}
\text{Attention}(\mathbf{Q},\mathbf{K},\mathbf{V}) 
= \text{softmax}\!\left(\frac{\mathbf{Q}\mathbf{K}^\top}{\sqrt{d_h}}\right)\mathbf{V},
\end{equation}
where $d_h$ is the embedding dimension of a single attention head.
Multi-head attention concatenates $H$ such projections to enhance expressivity~\cite{vaswani2017transformer}.
The bottleneck of this operation is the full attention matrix $\mathbf{A} = \mathbf{Q}\mathbf{K}^\top$, which grows quadratically with the number of input tokens.
For a moderate input length of $10$ frames at resolution $294\times 518$, the global attention matrix would already contain $(10\cdot \frac{294\cdot518}{14^2})^2\approx 1.2\cdot10^8$ elements, corresponding to more than 100MB at half precision.
At $1000$ frames, the required space would increase to more than 1 TB.

\PAR{Block-Sparse Attention.}
Sparse attention reduces the quadratic cost of full attention by constraining the pattern of non-zero entries in $\mathbf{Q}\mathbf{K}^\top$.
Instead of attending to all positions, only predefined entries are computed:
\begin{equation}
\text{SparseAttn}(\mathbf{Q},\mathbf{K},\mathbf{V}) 
= \text{softmax}\!\left(\frac{(\mathbf{Q}\mathbf{K}^\top) \odot \mathbf{M}}{\sqrt{d_h}}\right)\mathbf{V},
\end{equation}
where $\mathbf{M}$ is a binary sparsity mask and the $\odot$ operator denotes the element-wise (Hadamard) product.
In practice, block-sparse masks are preferred, as they align with modern hardware accelerators, enabling efficient memory access and parallelization \cite{child2019sparseTransformer,beltagy2020longformer,dao2022flashattention}.

\begin{figure}
    \centering
        \includegraphics{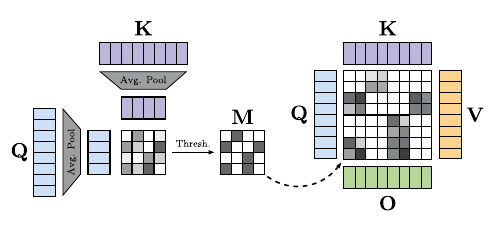}
    \caption{
    \textbf{Overview of the training-free adaptive sparse attention.}
    Keys and queries are average pooled to get a low-resolution approximation of the attention map.
    This low-resolution attention map is used to create the binary mask for block-sparse attention by Top-K selection and adaptive thresholding.
    }
    \label{fig:sparse-attention-overview}
\end{figure}

\PAR{Predicting Block Masks.}
To predict the importance of blocks from queries and keys, we first apply average pooling with block size $b$ to the queries $\mathbf{Q}$ and keys $\mathbf{K}$, yielding pooled representations $P^b(\mathbf{Q})$ and $P^b(\mathbf{K})$. 
We then compute their similarity matrix $S = P^b(\mathbf{Q}) P^b(\mathbf{K})^\top$ and apply a softmax to obtain probability distributions over blocks.
The resulting distributions provide a natural ranking of candidate blocks, from which we derive a binary block-sparsity mask using combined top-$k$ and cumulative density function (CDF) thresholding.
The mask directly determines the query--key blocks to evaluate, and can be used directly with standard block-sparse attention kernels.
A graphical overview is provided in~\cref{fig:sparse-attention-overview}.
We experimented with an additional linear projection layer after pooling queries and keys, but found no improvement over the baseline.

\PAR{Selecting Blocks.} 
We select blocks using two complementary criteria: a CDF threshold $\tau$ and a sparsity ratio $\rho$.
The CDF threshold ensures the selected set covers at least a $\tau$ fraction of predicted cumulative probability.
The sparsity ratio acts as a lower bound, enforcing a minimum of $k = \lfloor B \cdot (1 - \rho) \rfloor$ top-ranked blocks, where $B$ denotes the total number of blocks.
The two criteria complement each other: In uniform layers, a fixed sparsity ratio alone may admit too few blocks, while the CDF threshold guarantees sufficient coverage; in sparse layers, the threshold may be met with very few blocks, while the sparsity ratio ensures that at least a reasonable minimum is preserved. 
Together, they adapt block selection to diverse sparsity patterns while maintaining stability and efficiency.

\PAR{Comparison to SpargeAttention.}
While our method is related to SpargeAttention~\cite{zhang2025spargeattention}, the approaches differ in design and implementation trade-offs.
SpargeAttention introduces additional mechanisms such as self-similarity filtering and warp-level PV pruning, which increases implementation complexity and maintenance overhead across GPU generations.
In contrast, our method is kernel-agnostic.
It produces a binary block mask from pooled query--key similarities, and blocks are chosen based on a fixed sparsity ratio and a CDF threshold, hence compatible with any standard block-sparse attention kernel.
This decoupling from hardware-specific optimizations makes our approach simpler to implement, easier to maintain, and more robust to future accelerator iterations.

\subsection{Handling Special Tokens}
\label{ssec:special-token-handling}
VGGT employs register tokens~\cite{darcet2023register} in addition to learned camera tokens which differentiate between frames.
We find that these tokens behave qualitatively differently from the regular patch tokens, and pooling them together results in worse performance for similar sparsity ratios.
We therefore split the tokens into two sets: $\mathbf{X}_p$ containing the patch tokens, and $\mathbf{X}_s$, containing the special tokens, \ie register and camera tokens.
We only predict and use the block sparse map on the patch tokens $\mathbf{X}_p$, and compute the full attention on $\mathbf{X}_s$, as well as cross-attention between $\mathbf{X}_s$ and $\mathbf{X}_p$ and vice-versa.
We find that this strategy is essential to avoid large drops in task performance.

\section{Experiments}
\begin{figure*}
    \centering
    \begin{minipage}{0.5\textwidth}
        \begin{tikzpicture}

\pgfplotstableread[col sep=comma]{fig/rawdata/re10k/sparse_re10k_vggt.csv}\tableReTenKVGGT
\pgfplotstableread[col sep=comma]{fig/rawdata/re10k/sparse_re10k_pi3.csv}\tableReTenKPI
\pgfplotstableread[col sep=comma]{fig/rawdata/re10k/sparse_re10k_ma.csv}\tableReTenKMA

\pgfplotstableread[col sep=comma]{fig/rawdata/co3d/sparse_co3d_vggt.csv}\tableCOThreeDVGGT
\pgfplotstableread[col sep=comma]{fig/rawdata/co3d/sparse_co3d_pi3.csv}\tableCOThreeDPI
\pgfplotstableread[col sep=comma]{fig/rawdata/co3d/sparse_co3d_ma.csv}\tableCOThreeDMA

\begin{groupplot}[
    group style={
        columns=2,
        yticklabels at=edge left,
        horizontal sep=8pt,
    },
    axis lines=left,
    width=5cm,
    height=4cm,
    xlabel={Avg sparsity},
    ylabel={AUC@30 $\uparrow$},
    ymin=60, ymax=95,
    grid=both,
    grid style={gray!20},
    xmin=0, xmax=90,
    title style={yshift=-5pt},
]

\nextgroupplot[title={Real Estate 10K}]

\addplot[SVGGTMarker]  table[x=sparsity, y expr=\thisrow{AUC30}*100]{\tableReTenKVGGT};
\addplot[SPIMarker]  table[x=sparsity, y expr=\thisrow{AUC30}*100]{\tableReTenKPI};
\addplot[SMAMarker]  table[x=sparsity, y expr=\thisrow{AUC30}*100]{\tableReTenKMA};

\addplot[domain=0:100, VGGTMarker] {77.5}; %
\addplot[domain=0:100, PIMarker]  {87.4}; %
\addplot[domain=0:100, MAMarker] {78.2}; %
\addplot[domain=0:100, FasterMarker] {72.9}; %
\addplot[domain=0:100, CuterMarker]  {74.1}; %
\addplot[domain=0:100, FlareMarker] {78.8}; %

\nextgroupplot[title={CO3Dv2}, ylabel={}]

\addplot[SVGGTMarker]  table[x=sparsity, y expr=\thisrow{AUC30}*100]{\tableCOThreeDVGGT};
\addplot[SPIMarker]  table[x=sparsity, y expr=\thisrow{AUC30}*100]{\tableCOThreeDPI};
\addplot[SMAMarker]  table[x=sparsity, y expr=\thisrow{AUC30}*100]{\tableCOThreeDMA};

\addplot[domain=0:100, VGGTMarker] {90.9}; %
\addplot[domain=0:100, PIMarker]  {91.2}; %
\addplot[domain=0:100, MAMarker] {69.4}; %
\addplot[domain=0:100, FasterMarker] {83.6}; %
\addplot[domain=0:100, CuterMarker]  {74.1}; %
\addplot[domain=0:100, FlareMarker] {75.2}; %
\end{groupplot}
\end{tikzpicture}
    \end{minipage}%
    \begin{minipage}{0.5\textwidth}
        \begin{tikzpicture}

\pgfplotstableread[col sep=comma]{fig/rawdata/tum/sparse_tum_vggt.csv}\tableTUMVGGT
\pgfplotstableread[col sep=comma]{fig/rawdata/tum/sparse_tum_pi3.csv}\tableTUMPI
\pgfplotstableread[col sep=comma]{fig/rawdata/tum/sparse_tum_ma.csv}\tableTUMMA

\pgfplotstableread[col sep=comma]{fig/rawdata/scannet/sparse_scannet_vggt.csv}\tableScanNetVGGT
\pgfplotstableread[col sep=comma]{fig/rawdata/scannet/sparse_scannet_pi3.csv}\tableScanNetPI
\pgfplotstableread[col sep=comma]{fig/rawdata/scannet/sparse_scannet_ma.csv}\tableScanNetMA

\begin{groupplot}[
    group style={
        columns=2,
        yticklabels at=edge left,
        horizontal sep=8pt,
    },
    axis lines=left,
    width=5cm,
    height=4cm,
    xlabel={Avg sparsity},
    ylabel={ATE $\downarrow$},
    grid=both,
    grid style={gray!20},
    xmin=0, xmax=90,
    ymin=0.004, ymax=0.11,
    y tick label style={
        /pgf/number format/.cd,
        fixed,
        fixed zerofill,
        precision=2,
        /tikz/.cd
    },
    title style={yshift=-5pt},
]
    
\nextgroupplot[title={TUM}, ylabel={ATE $\downarrow$}]

\addplot[SVGGTMarker]  table[x=sparsity, y=ATE]{\tableTUMVGGT};
\addplot[SPIMarker]  table[x=sparsity, y=ATE]{\tableTUMPI};
\addplot[SMAMarker]  table[x=sparsity, y=ATE]{\tableTUMMA};

\addplot[domain=0:100, VGGTMarker] {0.012}; %
\addplot[domain=0:100, PIMarker]  {0.014}; %
\addplot[domain=0:100, MAMarker] {0.026}; %
\addplot[domain=0:100, FasterMarker] {0.044}; %
\addplot[domain=0:100, CuterMarker]  {0.043}; %
\addplot[domain=0:100, FlareMarker] {0.023}; %

\nextgroupplot[title={ScanNet}, ylabel={}]

\addplot[SVGGTMarker]  table[x=sparsity, y=ATE]{\tableScanNetVGGT};
\addplot[SPIMarker]  table[x=sparsity, y=ATE]{\tableScanNetPI};
\addplot[SMAMarker]  table[x=sparsity, y=ATE]{\tableScanNetMA};

\addplot[domain=0:100, VGGTMarker] {0.038}; %
\addplot[domain=0:100, PIMarker]  {0.034}; %
\addplot[domain=0:100, MAMarker] {0.060}; %
\addplot[domain=0:100, FasterMarker] {0.091}; %
\addplot[domain=0:100, CuterMarker]  {0.113}; %
\addplot[domain=0:100, FlareMarker] {0.061}; %

\end{groupplot}
\end{tikzpicture}
    \end{minipage}
    \begin{tikzpicture}

\pgfplotstableread[col sep=comma]{fig/rawdata/7scenes/sparse_7scenes_vggt.csv}\tableSevenScenesVGGT
\pgfplotstableread[col sep=comma]{fig/rawdata/7scenes/sparse_7scenes_pi3.csv}\tableSevenScenesPI
\pgfplotstableread[col sep=comma]{fig/rawdata/7scenes/sparse_7scenes_ma.csv}\tableSevenScenesMA

\pgfplotstableread[col sep=comma]{fig/rawdata/nrgbd/sparse_nrgbd_vggt.csv}\tableNRGBDVGGT
\pgfplotstableread[col sep=comma]{fig/rawdata/nrgbd/sparse_nrgbd_pi3.csv}\tableNRGBDPI
\pgfplotstableread[col sep=comma]{fig/rawdata/nrgbd/sparse_nrgbd_ma.csv}\tableNRGBDMA

\pgfplotstableread[col sep=comma]{fig/rawdata/dtu/sparse_dtu_vggt.csv}\tableDTUVGGT
\pgfplotstableread[col sep=comma]{fig/rawdata/dtu/sparse_dtu_pi3.csv}\tableDTUPI
\pgfplotstableread[col sep=comma]{fig/rawdata/dtu/sparse_dtu_ma.csv}\tableDTUMA

\pgfplotstableread[col sep=comma]{fig/rawdata/eth3d/sparse_eth3d_vggt.csv}\tableETHVGGT
\pgfplotstableread[col sep=comma]{fig/rawdata/eth3d/sparse_eth3d_pi3.csv}\tableETHPI
\pgfplotstableread[col sep=comma]{fig/rawdata/eth3d/sparse_eth3d_ma.csv}\tableETHMA

\begin{groupplot}[
    group style={
        columns=4,
        horizontal sep=0.83cm,
    },
    axis lines=left,
    width=4.9cm,
    height=4cm,
    xlabel={Avg sparsity},
    ylabel={Chamfer dist. $\downarrow$},
    grid=both,
    grid style={gray!20},
    xmin=0, xmax=88,
    scaled y ticks=false,
    tick label style={/pgf/number format/fixed},
    legend style={
        draw=none,
        at={(-0.5, -0.5)},
        anchor=north,
        legend columns=3,
        font=\small,
        /tikz/every even column/.append style={column sep=3pt}
    },
    legend to name=regressionMVReconLegend,
    title style={yshift=-5pt},
]

\nextgroupplot[title={7Scences}, ymin=0.02, ymax=0.2]
\coordinate (c1) at (rel axis cs:0,1);
\addplot[SVGGTMarker]  table[x=sparsity, y=cham]{\tableSevenScenesVGGT};
\addplot[SPIMarker]  table[x=sparsity, y=cham]{\tableSevenScenesPI};
\addplot[SMAMarker]  table[x=sparsity, y=cham]{\tableSevenScenesMA};

\addplot[domain=0:100, VGGTMarker] {0.061}; %
\addplot[domain=0:100, PIMarker]  {0.060}; %
\addplot[domain=0:100, MAMarker] {0.083}; %
\addplot[domain=0:100, FasterMarker] {0.136}; %
\addplot[domain=0:100, CuterMarker]  {0.101}; %
\addplot[domain=0:100, FlareMarker] {0.109}; %

\nextgroupplot[title={NRGBD}, ylabel={}, ymin=0.02, ymax=0.2]
\addplot[SVGGTMarker]  table[x=sparsity, y=cham]{\tableNRGBDVGGT};
\addplot[SPIMarker]  table[x=sparsity, y=cham]{\tableNRGBDPI};
\addplot[SMAMarker]  table[x=sparsity, y=cham]{\tableNRGBDMA};

\addplot[domain=0:100, VGGTMarker] {0.049}; %
\addplot[domain=0:100, PIMarker]  {0.028}; %
\addplot[domain=0:100, MAMarker] {0.093}; %
\addplot[domain=0:100, FasterMarker] {0.135}; %
\addplot[domain=0:100, CuterMarker]  {0.103}; %
\addplot[domain=0:100, FlareMarker] {0.062}; %

\nextgroupplot[title={DTU}, ylabel={}, ymin=1.0, ymax=3.5]
\addplot[SVGGTMarker]  table[x=sparsity, y=cham]{\tableDTUVGGT};
\addplot[SPIMarker]  table[x=sparsity, y=cham]{\tableDTUPI};
\addplot[SMAMarker]  table[x=sparsity, y=cham]{\tableDTUMA};

\addplot[domain=0:100, VGGTMarker] {1.20}; %
\addplot[domain=0:100, PIMarker]  {1.50}; %
\addplot[domain=0:100, MAMarker] {2.45}; %
\addplot[domain=0:100, FasterMarker] {3.32}; %
\addplot[domain=0:100, CuterMarker]  {24.94}; %

\nextgroupplot[title={ETH3D}, ylabel={}, ymin=0.08, ymax=0.8]
\addplot[SVGGTMarker]  table[x=sparsity, y=cham]{\tableETHVGGT};
\addlegendentry{Sparse VGGT (Ours)}
\addplot[SPIMarker]  table[x=sparsity, y=cham]{\tableETHPI};
\addlegendentry{Sparse $\pi^3$ (Ours)}
\addplot[SMAMarker]  table[x=sparsity, y=cham]{\tableETHMA};
\addlegendentry{Sparse MapAnything (Ours)}

\addplot[domain=0:100, VGGTMarker] {0.265}; %
\addlegendentry{VGGT~\cite{wang2025vggt}}
\addplot[domain=0:100, PIMarker]  {0.099}; %
\addlegendentry{$\pi^3$~\cite{wang2025pi}}
\addplot[domain=0:100, MAMarker] {0.194}; %
\addlegendentry{MapAnything~\cite{keetha2025mapanything}}
\addplot[domain=0:100, FasterMarker] {0.551}; %
\addlegendentry{Fast3r~\cite{yang2025fast3r}}
\addplot[domain=0:100, CuterMarker]  {0.585}; %
\addlegendentry{CUT3R~\cite{wang2025cut3r}}
\addplot[domain=0:100, FlareMarker] {0.465}; %
\addlegendentry{FLARE~\cite{zhang2025flare}}

\coordinate (c2) at (rel axis cs:1,1);
\end{groupplot}

\coordinate (c3) at ($(c1)!.5!(c2)$);
    \node[below=-1mm] at (c3 |- current bounding box.south) {\pgfplotslegendfromname{regressionMVReconLegend}};
\end{tikzpicture}
    \caption{
        \textbf{Results for Relative Pose Estimation (top) Multi-View Reconstruction (bottom).}
        Multi-view reconstruction performance seems to be robust against sparsification of global attention; even in the highest sparsity settings, the results are on par or better than other state-of-the-art methods.
        We provide comprehensive tables for these results in the supplementary material.
    }
    \label{fig:sparsity-regression}
\end{figure*}

\begin{figure*}
    \centering
    \input{fig/qualitative_samples.tex}
    \caption{
        \textbf{Qualitative examples.}
        We show examples from the ETH3D dataset~\cite{schoeps2017eth3d}.
        Increasing sparsity leads to small perturbations in the reconstruction, but the overall quality stays remarkably high.
        Additional qualitative examples can be found in the supplementary material.
    }
    \label{fig:qualitative}
\end{figure*}

We extend three large reconstruction models, VGGT~\cite{wang2025vggt}, $\pi^3$~\cite{wang2025pi}, and MapAnything~\cite{keetha2025mapanything} with the described sparse global attention mechanism and evaluate the performance impact on common multi-view benchmarks.
In particular, we evaluate the robustness of the models against varying \emph{effective sparsity ratios}, which we define as the ratio of computed entries of the attention map to the total number of entries.
As the number of input frames increases, the theoretical speed-up from a sparsity ratio of $x$ approaches $\frac{1}{1 - x}$, since the overall computational costs are dominated by the global attention.
Processing a $200$ frame sequence with sparsity ratio of $0.75$, for example, will accelerate the global attention by a factor of four.
Depending on the ratio of global attention to the overall runtime, the end-to-end speed-up will be lower.
A comprehensive analysis of the achievable speed-up is provided in the supplementary.

In the main text, we show results for common benchmarks on relative pose estimation (Real Estate 10K~\cite{zhou2018re10k}, Common Objects in 3D~\cite{reizenstein21co3d} TUM~\cite{sturm2012tumdynamic}, and ScanNet~\cite{dai2017scannet}), and pointmap estimation (7Scenes~\cite{shotton2013sevenscenes}, NRGBD~\cite{azinovic2022nrgbd}, DTU~\cite{jensen2014dtu}, and ETH3D~\cite{schoeps2017eth3d}).
Additionally, we show results for scene-level pose estimation on the Tanks \& Temples dataset~\cite{knapitsch2017tanksAndTemples}.
In the supplementary materials, we also show results on longer sequences of the ScanNet~\cite{dai2017scannet} dataset.

\subsection{Regression Tests}
We evaluate our sparsified model on relative pose and pointmap estimation, using a range of common benchmark datasets.
Unless indicated otherwise, we closely follow the setting of the original VGGT paper~\cite{wang2025vggt}.
For a comprehensive high-level overview of the results, we visualize the sparsity-performance trade-off in~\cref{fig:sparsity-regression}.
The results show that task performance degrades comparatively little with increasing sparsity.
Both VGGT and $\pi^3$ achieve results comparable to other state-of-the-art models even at high sparsity ratios, which is further corroborated by the qualitative examples provided in \cref{fig:qualitative}.

\PAR{Pose Estimation} involves determining the 6DoF pose for each input view.
We evaluate AUC@30 on Common Objects in 3D~\cite{reizenstein21co3d} and Real Estate 10K~\cite{zhou2018re10k}, as well as ATE on TUM Dynamic~\cite{sturm2012tumdynamic} and ScanNet~\cite{dai2017scannet}.
For these experiments on TUM and ScanNet, we follow the setup of previous works~\cite{zhang2025monst3r,wang2025pi} and sample 90 frames with a temporal stride of 3 from the beginning of each sequence.
We show aggregate results in~\cref{fig:sparsity-regression} (top).
While we observe a continuous decrease in pose accuracy for increasing sparsity levels, the modified model still performs comparable to other state-of-the-art methods.

\PAR{Multi-view Pointmap Estimation} is the task of predicting the surface 3D point for each pixel in all input images.
We evaluate on 7Scenes~\cite{shotton2013sevenscenes}, NRGBD~\cite{azinovic2022nrgbd}, ETH3D~\cite{schoeps2017eth3d}, and DTU~\cite{jensen2014dtu} following $\pi^3$~\cite{wang2025pi} and VGGT~\cite{wang2025vggt}.
We show a compact representation of the results in~\cref{fig:sparsity-regression} and full result tables in the supplementary.
We observe similar trends as before: task performance stays at an acceptable level even at high sparsity ratios.
The apparent improvements on ETH3D are likely attributable to randomness.
Overall, we do not expect the sparse models to outperform the baseline.

\subsection{Long-Sequence Relative Pose Estimation}
To test the scalability of our block-sparse attention scheme, we evaluate pose estimation on the Tanks \& Temples benchmark~\cite{knapitsch2017tanksAndTemples}.
We report AUC@30 in~\cref{fig:regression-tandt}, for which we observe very little degradation with increasing effective sparsity.
On sequences with 200 frames, our sparse $\pi^3$ model with an effective sparsity ratio of 75\% runs around twice as fast as the baseline model on an H100 GPU; on longer sequences, the speed-up increases further.
An analysis of the achievable speed-up, time measurements, as well as similar experiments on long sequences from the ScanNet~\cite{dai2017scannet} dataset can be found in the supplementary materials.

\begin{figure}
\centering
\begin{tikzpicture}

\pgfplotstableread[col sep=comma]{fig/rawdata/tandt/sparse_tandt_200_vggt.csv}\tableTTTwoHundredVGGT
\pgfplotstableread[col sep=comma]{fig/rawdata/tandt/sparse_tandt_200_pi3.csv}\tableTTTwoHundredPI
\pgfplotstableread[col sep=comma]{fig/rawdata/tandt/sparse_tandt_200_ma.csv}\tableTTTwoHundredMA

\pgfplotstableread[col sep=comma]{fig/rawdata/tandt/sparse_tandt_100_vggt.csv}\tableTTOneHundredVGGT
\pgfplotstableread[col sep=comma]{fig/rawdata/tandt/sparse_tandt_100_pi3.csv}\tableTTOneHundredPI
\pgfplotstableread[col sep=comma]{fig/rawdata/tandt/sparse_tandt_100_ma.csv}\tableTTOneHundredMA

\pgfplotstableread[col sep=comma]{fig/rawdata/tandt/sparse_tandt_50_vggt.csv}\tableTTFiftyVGGT
\pgfplotstableread[col sep=comma]{fig/rawdata/tandt/sparse_tandt_50_pi3.csv}\tableTTFiftyPI
\pgfplotstableread[col sep=comma]{fig/rawdata/tandt/sparse_tandt_50_ma.csv}\tableTTFiftyMA

\begin{groupplot}[
    group style={
        rows=3,
        xticklabels at=edge bottom,
        vertical sep=5pt,
    },
    width=8cm,
    height=3.2cm,
    axis lines=left,
    xlabel={Avg sparsity},
    ylabel={AUC@30$\uparrow$},
    grid=both,
    grid style={gray!20},
    xmin=0, xmax=90,
    ymin=75, ymax=99,
    scaled y ticks=false,
    tick label style={/pgf/number format/fixed},
    legend style={
        draw=none,
        at={(-0.5, -0.5)},
        anchor=north,
        legend columns=5,
        font=\small,
        /tikz/every even column/.append style={column sep=3pt}
    },
    legend to name=TandtLegend,
]

\nextgroupplot[ylabel={}, xlabel={}]
\node[anchor=north east] at (rel axis cs:1,1) {50 Frames};
\addplot[SVGGTMarker]  table[x=sparsity, y expr=\thisrow{AUC30}*100]{\tableTTFiftyVGGT};
\addplot[SPIMarker]   table[x=sparsity, y expr=\thisrow{AUC30}*100]{\tableTTFiftyPI};
\addplot[SMAMarker]   table[x=sparsity, y expr=\thisrow{AUC30}*100]{\tableTTFiftyMA};

\nextgroupplot[xlabel={}]
\node[anchor=north east] at (rel axis cs:1,1) {100 Frames};
\addplot[SVGGTMarker]  table[x=sparsity, y expr=\thisrow{AUC30}*100]{\tableTTOneHundredVGGT};
\addplot[SPIMarker]   table[x=sparsity, y expr=\thisrow{AUC30}*100]{\tableTTOneHundredPI};
\addplot[SMAMarker]   table[x=sparsity, y expr=\thisrow{AUC30}*100]{\tableTTOneHundredMA};

\nextgroupplot[ylabel={}]
\node[anchor=north east] at (rel axis cs:1,1) {200 Frames};
\addplot[SVGGTMarker]  table[x=sparsity, y expr=\thisrow{AUC30}*100]{\tableTTTwoHundredVGGT};
\addlegendentry{Sparse VGGT}
\addplot[SPIMarker]   table[x=sparsity, y expr=\thisrow{AUC30}*100]{\tableTTTwoHundredPI};
\addlegendentry{Sparse $\pi^3$}
\addplot[SMAMarker]   table[x=sparsity, y expr=\thisrow{AUC30}*100]{\tableTTTwoHundredMA};
\addlegendentry{Sparse MapAnything}

\coordinate (c2) at (rel axis cs:1,1);
\end{groupplot}

\coordinate (c3) at ($(c1)!.5!(c2)$);
    \node[below] at (c3 |- current bounding box.south) {\pgfplotslegendfromname{TandtLegend}};
\end{tikzpicture}
\caption{
\textbf{Pose estimation on Tanks \& Temples for different input sizes and sparsity ratios.}
We show more comprehensive results in the supplementary materials.
}
\label{fig:regression-tandt}
\end{figure}

\begin{figure}
    \centering
    \begin{tikzpicture}

\pgfplotstableread[col sep=comma]{fig/rawdata/co3d/ablation_co3d_ours.csv}\tableCOThreeDOurs
\pgfplotstableread[col sep=comma]{fig/rawdata/co3d/ablation_co3d_ours_no_special.csv}\tableCOThreeDOursNoSpecial
\pgfplotstableread[col sep=comma]{fig/rawdata/co3d/ablation_co3d_ours_no_topk.csv}\tableCOThreeDOursNoTopk
\pgfplotstableread[col sep=comma]{fig/rawdata/co3d/ablation_co3d_cdf_only.csv}\tableCOThreeDCDFOnly
\pgfplotstableread[col sep=comma]{fig/rawdata/co3d/ablation_co3d_sparge.csv}\tableCOThreeDSparge
\pgfplotstableread[col sep=comma]{fig/rawdata/co3d/ablation_co3d_sparge_hsim.csv}\tableCOThreeDSpargeHighSim
\pgfplotstableread[col sep=comma]{fig/rawdata/co3d/ablation_co3d_topk_only.csv}\tableCOThreeDTopkOnly
\pgfplotstableread[col sep=comma]{fig/rawdata/co3d/ablation_co3d_random_no_special.csv}\tableCOThreeDRandomNoSpecial
\pgfplotstableread[col sep=comma]{fig/rawdata/co3d/ablation_co3d_random_patch.csv}\tableCOThreeDRandomPatch

\pgfplotstableread[col sep=comma]{fig/rawdata/eth3d/ablation_eth3d_ours.csv}\tableETHThreeDOurs
\pgfplotstableread[col sep=comma]{fig/rawdata/eth3d/ablation_eth3d_ours_no_special.csv}\tableETHThreeDOursNoSpecial
\pgfplotstableread[col sep=comma]{fig/rawdata/eth3d/ablation_eth3d_ours_no_topk.csv}\tableETHThreeDOursNoTopk
\pgfplotstableread[col sep=comma]{fig/rawdata/eth3d/ablation_eth3d_cdf_only.csv}\tableETHThreeDCDFOnly
\pgfplotstableread[col sep=comma]{fig/rawdata/eth3d/ablation_eth3d_sparge.csv}\tableETHThreeDSparge
\pgfplotstableread[col sep=comma]{fig/rawdata/eth3d/ablation_eth3d_sparge_hsim.csv}\tableETHThreeDSpargeHighSim
\pgfplotstableread[col sep=comma]{fig/rawdata/eth3d/ablation_eth3d_topk_only.csv}\tableETHThreeDTopkOnly
\pgfplotstableread[col sep=comma]{fig/rawdata/eth3d/ablation_eth3d_random_no_special.csv}\tableETHThreeDRandomNoSpecial
\pgfplotstableread[col sep=comma]{fig/rawdata/eth3d/ablation_eth3d_random_patch.csv}\tableETHThreeDRandomPatch

\begin{groupplot}[
    group style={
        rows=2,
        xticklabels at=edge bottom,
        vertical sep=10pt,
    },
    width=8.3cm,
    height=3.9cm,
    axis lines=left,
    xlabel={Avg sparsity},
    grid=both,
    grid style={gray!20},
    xmin=0, xmax=90,
    ymin=0.4, ymax=0.9,
    scaled y ticks=false,
    tick label style={/pgf/number format/fixed},
    legend style={
        draw=none,
        at={(-0.5, -0.5)},
        anchor=north,
        legend columns=2,
        font=\small,
        /tikz/every even column/.append style={column sep=3pt}
    },
    legend to name=AblationsLegend,
]

\nextgroupplot[ylabel={{AUC@30$\uparrow$}}, xlabel={}, ymin=10, ymax=99]
\addplot[only marks, thick, mark=x, blue]  table[x=sparsity, y expr=\thisrow{AUC30}*100]{\tableCOThreeDOurs};
\addplot[only marks, thick, mark=x, black]  table[x=sparsity, y expr=\thisrow{AUC30}*100]{\tableCOThreeDOursNoSpecial};
\addplot[only marks, thick, mark=star, red]  table[x=sparsity, y expr=\thisrow{AUC30}*100]{\tableCOThreeDOursNoTopk};
\addplot[only marks, thick, mark=star, orange]  table[x=sparsity, y expr=\thisrow{AUC30}*100]{\tableCOThreeDCDFOnly};
\addplot[only marks, thick, mark=diamond*, green!80!black]  table[x=sparsity, y expr=\thisrow{AUC30}*100]{\tableCOThreeDSparge};
\addplot[only marks, thick, mark=diamond*, gray]  table[x=sparsity, y expr=\thisrow{AUC30}*100]{\tableCOThreeDSpargeHighSim};
\addplot[only marks, thick, mark=o, blue]  table[x=sparsity, y expr=\thisrow{AUC30}*100]{\tableCOThreeDRandomNoSpecial};
\addplot[only marks, thick, mark=o, black]  table[x=sparsity, y expr=\thisrow{AUC30}*100]{\tableCOThreeDRandomPatch};

\node[anchor=south west] at (rel axis cs:0,0) {CO3Dv2};

\nextgroupplot[ylabel={{Chamfer dist.$\downarrow$}}, ymin=0.1, ymax=0.7]
\coordinate (c1) at (rel axis cs:0,1);

\addplot[only marks, thick, mark=x, blue]  table[x=sparsity, y expr=\thisrow{cham}]{\tableETHThreeDOurs};
\addlegendentry{Ours}
\addplot[only marks, thick, mark=x, black]  table[x=sparsity, y expr=\thisrow{cham}]{\tableETHThreeDOursNoSpecial};
\addlegendentry{Ours w/o Dense Special}
\addplot[only marks, thick, mark=star, red]  table[x=sparsity, y expr=\thisrow{cham}]{\tableETHThreeDOursNoTopk};
\addlegendentry{CDF only w/ dense special}
\addplot[only marks, thick, mark=star, orange]  table[x=sparsity, y expr=\thisrow{cham}]{\tableETHThreeDCDFOnly};
\addlegendentry{CDF only w/o dense special}
\addplot[only marks, thick, mark=diamond*, green!80!black]  table[x=sparsity, y expr=\thisrow{cham}]{\tableETHThreeDSparge};
\addlegendentry{SpargeAttn}
\addplot[only marks, thick, mark=diamond*, gray]  table[x=sparsity, y expr=\thisrow{cham}]{\tableETHThreeDSpargeHighSim};
\addlegendentry{SpargeAttn w/ high sim thr.}
\addplot[only marks, thick, mark=o, blue]  table[x=sparsity, y expr=\thisrow{cham}]{\tableETHThreeDRandomNoSpecial};
\addlegendentry{Random}
\addplot[only marks, thick, mark=o, black]  table[x=sparsity, y expr=\thisrow{cham}]{\tableETHThreeDRandomPatch};
\addlegendentry{Random w/ dense special}

\node[anchor=north west] at (rel axis cs:0,1) {ETH3D};

\coordinate (c2) at (rel axis cs:1,1);
\end{groupplot}
\coordinate (c3) at ($(c1)!.425!(c2)$);
    \node[below] at (c3 |- current bounding box.south) {\pgfplotslegendfromname{AblationsLegend}};
\end{tikzpicture}
    \caption{
        \textbf{Baseline sparse attention comparison.}
        All shown experiments do not employ SpargeAttention's hilbert permutation.
        See the appendix for a more comprehensive comparison.
    }
    \label{fig:ablation-baselines}
\end{figure}

\subsection{Ablations}
\label{ssec:ablations-main}
To validate our design decisions, we compare our method to SpargeAttention~\cite{zhang2025spargeattention} and a random masking baseline in~\cref{fig:ablation-baselines}; we provide additional results in the supplementary material, including a comparison to SeerAttention~\cite{gao2024seerattention}.
As expected, performance with random masking drops rapidly as the sparsity ratio increases.
SpargeAttention utilizes CDF-thresholding as the main sparsity mechanism and keeps additional tokens based on intra-block similarity thresholding; for low sparsity ratios, their performance is similar to ours, but with increasing sparsity performance drops.
Since our method combines Top-K and CDF thresholding, it is more robust against high sparsity ratios.

\section{Discussion}
We analyzed global attention in transformer-based geometry estimators, VGGT and $\pi^3$, and found that it exhibits unstructured sparsity patterns, which can be interpreted as exhaustive correspondence search and is most pronounced in the middle aggregator layers.
Building on these observations, we adapted a block-sparse global attention mechanism for patch-patch interactions, improving scalability to large image sets.
Our approach achieves task performance comparable to the original model while being more than three times faster during inference on large-scale scenes.

While we focused on increasing the number of input images in this study, our approach is also applicable to speeding up inference on higher-resolution inputs.
Doubling the input resolution results in four times as many patch tokens and in turn to $16\times$ more compute spent on global attention, drastically exacerbating the existing bottleneck.

Exploiting block-wise sparsity is orthogonal to other acceleration approaches like FlashAttention~\cite{dao2022flashattention,dao2023flashattentionV2,shah2024flashattentionV3}, and can potentially be integrated into the training procedure to reduce the impact on task performance.

\small
\PAR{Acknowledgments.}
This research was supported, in parts, by BMFTR project bridgingAI (16DHBKI023), by the Bosch-RWTH Lighthouse Collaboration "Context Understanding for Autonomous Systems", and by the Robotics Institute Germany (RIG).
Computations were performed with computing resources granted by RWTH Aachen University under projects \verb|thes2009| and \verb|rwth1849|.
The authors thank Karim Knaebel and Kadir Yilmaz for helpful feedback and Laura Schneider for help with visualizations.

{
    \small
    \bibliographystyle{ieeenat_fullname}
    \bibliography{main}
}
\clearpage
\setcounter{page}{1}
\maketitlesupplementary

\appendix
\renewcommand{\thefigure}{A-\arabic{figure}}
\renewcommand{\thetable}{A-\arabic{table}}
\setcounter{figure}{0}
\setcounter{table}{0}

\section{Theoretical FLOPs Reduction}
\label{ssec:flops}
We derive the number of floating point operations (FLOPs) that is required for a forward pass of a model with block-sparse attention using a sparsity ratio of $\rho < 1$ and the orginal VGGT model with dense attention ($\rho = 1$).
The ratio between these two numbers serves as an upper bound of the achievable speed-up for a given sparsity ratio.
We take only the encoder and the aggregator into account for these calculations.
Following common practice, we only consider FLOPs of matrix multiplications in the FFN and Attention layers, counting a multiply-add as a single instruction.
We ignore operations
related to patchification, embedding layers, register tokens, and task-specific output heads.

We base our calculations on a model using a per-frame transformer encoder $E$ with model dimension $d_E$ and $L_E$ transformer blocks.
The aggregator is a transformer with model dimension $d_A$, employing $2L_A$ transformer blocks alternating between frame-wise and global attention.
The number of FLOPs needed to apply an MLP layer to a single tokens is
\begin{equation}
    \text{FLOP}_{\text{MLP}}(d) = d \cdot d_{\text{FF}} + d_{\text{FF}} \cdot d = 8d^2,
\end{equation}
where $d_{\text{FF}}$ is the intermediate dimension of the MLP; for the models discussed in this paper, $d_{\text{FF}} = 4d$.
The number of FLOPs for a self-attention layer consists of the FLOPs necessary for the linear projections and the computation of the query-key-value products.
Block-sparse attention reduces the FLOPs required for computing the query-key-value interactions by the sparsity factor $\rho$.
For $t$ tokens and model dimension $d$, this gives
\begin{equation}
    \text{FLOP}_{\text{Attn}}(t, d, \rho) = 4td^2 + (1 - \rho) 2t^2d.
\end{equation}
The flops required for computing a single transformer block (consisting of either frame-wise or global attention and an MLP layer) are then
\begin{equation}
    \text{FLOP}_{\text{Block}}(t, d) = \text{FLOP}_{\text{Attn}}(t, d, 0) + T\cdot\text{FLOP}_{\text{MLP}}(d).
\end{equation}
A VGGT-like model contains $L_F$ frame-wise transformer blocks (including the encoder layers) and $L_G$ global attention blocks, such that we can compute the total number of FLOPs as
\begin{equation}
    \text{FLOP}(N, T, \rho) = L_FN\cdot\text{FLOP}_{\text{B}}(T, d) + \text{FLOP}_{\text{B}}(NT, d, \rho).
\end{equation}

Based on these notations, we compute the theoretical speedup as
\begin{equation}
    \text{speedup} = \frac{\text{FLOP}(N, T, 0)}{\text{FLOP}(N, T, \rho)}.
\end{equation}
We show the theoretical speed-up for several configurations in~\cref{tab:speedup-theoretical}.

\begin{table}
    \centering
    
    \begin{tabular}{c|cccc}
        \toprule
         & \multicolumn{4}{c}{Sparsity ratio $\rho$} \\
        $N$  & 0.25 & 0.50 & 0.75 & 0.90 \\
        \midrule
        100  & 1.2 & 1.7 & 2.6 & 4.0 \\
        300  & 1.3 & 1.8 & 3.3 & 6.4 \\
        500  & 1.3 & 1.9 & 3.5 & 7.4 \\
        1000 & 1.3 & 1.9 & 3.7 & 8.5 \\
        \bottomrule
    \end{tabular}        
    \caption{
        \textbf{Theoretical end-to-end speed-up} of VGGT using block-sparse global attention \vs dense global attention, at different number of frames $N$ and sparsity ratios $\rho$.
        Assuming a resolution of $392\times 518$, corresponding to $T=28\times37 = 1036$ tokens/frame.
    }
    \label{tab:speedup-theoretical}
\end{table}

\section{Long-Sequence Experiments}
To validate that our method scales to longer input sequences, we evaluate the performance of block-sparse global attention models on complete sequences from the ScanNet~\cite{dai2017scannet} test set.
In order to keep compute demands reasonable, we evaluate on every second test set sequence, totalling 50 sequences.
Instead of taking the first $N$ frames with a fixed temporal stride (as in the main text), here we sample $N$ frames evenly spaced over the whole trajectory.
This sampling scheme increases the task difficulty considerably: not only is the number of input frames much larger, the trajectories are also significantly longer and more complex.
We show results for 100, 300, 500, and 1000 input frames in~\cref{fig:sup:regression-sn-long}.
For 100 input frames, VGGT and MapAnything reach an ATE of 0.18 and 0.24, respectively.
With increasing input sequence length, their task performance drops slightly.
$\pi^3$ shows the best results, achieving an ATE of 0.15 regardless of the number of input frames.

Our sparse variants keep similar task performance up to 60\% effective sparsity, resulting in a 1.5$\times$ end-to-end speed-up at 100 input frames and 2$\times$ at more than 300 frames.
At around 75\% effective sparsity rate, the task performance starts to degrade, but is still comparable to or better than other state-of-the-art models.

\begin{figure*}
    \centering
    \begin{tikzpicture}

\pgfplotstableread[col sep=comma]{fig/rawdata/scannet50/sparse_scannet50_1000_vggt.csv}\tableSNFiftyOneThousandVGGT
\pgfplotstableread[col sep=comma]{fig/rawdata/scannet50/sparse_scannet50_1000_pi3.csv}\tableSNFiftyOneThousandPI
\pgfplotstableread[col sep=comma]{fig/rawdata/scannet50/sparse_scannet50_1000_ma.csv}\tableSNFiftyOneThousandMA

\pgfplotstableread[col sep=comma]{fig/rawdata/scannet50/sparse_scannet50_500_vggt.csv}\tableSNFiftyFiveHundredVGGT
\pgfplotstableread[col sep=comma]{fig/rawdata/scannet50/sparse_scannet50_500_pi3.csv}\tableSNFiftyFiveHundredPI
\pgfplotstableread[col sep=comma]{fig/rawdata/scannet50/sparse_scannet50_500_ma.csv}\tableSNFiftyFiveHundredMA

\pgfplotstableread[col sep=comma]{fig/rawdata/scannet50/sparse_scannet50_300_vggt.csv}\tableSNFiftyThreeHundredVGGT
\pgfplotstableread[col sep=comma]{fig/rawdata/scannet50/sparse_scannet50_300_pi3.csv}\tableSNFiftyThreeHundredPI
\pgfplotstableread[col sep=comma]{fig/rawdata/scannet50/sparse_scannet50_300_ma.csv}\tableSNFiftyThreeHundredMA

\pgfplotstableread[col sep=comma]{fig/rawdata/scannet50/sparse_scannet50_100_vggt.csv}\tableSNFiftyOneHundredVGGT
\pgfplotstableread[col sep=comma]{fig/rawdata/scannet50/sparse_scannet50_100_pi3.csv}\tableSNFiftyOneHundredPI
\pgfplotstableread[col sep=comma]{fig/rawdata/scannet50/sparse_scannet50_100_ma.csv}\tableSNFiftyOneHundredMA

\begin{groupplot}[
    group style={
        group size=4 by 2,
        yticklabels at=edge left,
        xticklabels at=edge bottom,
        horizontal sep=8pt,
        vertical sep=8pt,
    },
    width=5.4cm,
    height=4.0cm,
    axis lines=left,
    xlabel={Avg sparsity},
    grid=both,
    grid style={gray!20},
    xmin=0, xmax=90,
    ymin=0.1, ymax=0.35,
    scaled y ticks=false,
    tick label style={/pgf/number format/fixed},
    legend style={
        draw=none,
        at={(-0.5, -0.5)},
        anchor=north,
        legend columns=5,
        font=\small,
        /tikz/every even column/.append style={column sep=3pt}
    },
    legend to name=SNLongLegend,
]

\nextgroupplot[ylabel={ATE$\downarrow$}, xlabel={}]
\coordinate (c2) at (rel axis cs:0,0);
\node[anchor=north west] at (rel axis cs:0,1) {100 Frames};
\addplot[SVGGTMarker]  table[x=sparsity, y expr=\thisrow{ATE}]{\tableSNFiftyOneHundredVGGT};
\addplot[SPIMarker]   table[x=sparsity, y expr=\thisrow{ATE}]{\tableSNFiftyOneHundredPI};
\addplot[SMAMarker]   table[x=sparsity, y expr=\thisrow{ATE}]{\tableSNFiftyOneHundredMA};
\addplot[domain=0:100, VGGTMarker] {0.184}; %
\addplot[domain=0:100, PIMarker]  {0.151}; %
\addplot[domain=0:100, MAMarker] {0.242}; %
\addplot[domain=0:100, FasterMarker] {0.755}; %
\addplot[domain=0:100, CuterMarker]  {0.355}; %
\addplot[domain=0:100, FlareMarker] {0.430}; %

\nextgroupplot[ylabel={}, xlabel={}]
\node[anchor=north west] at (rel axis cs:0,1) {300 Frames};
\addplot[SVGGTMarker]  table[x=sparsity, y expr=\thisrow{ATE}]{\tableSNFiftyThreeHundredVGGT};
\addplot[SPIMarker]   table[x=sparsity, y expr=\thisrow{ATE}]{\tableSNFiftyThreeHundredPI};
\addplot[SMAMarker]   table[x=sparsity, y expr=\thisrow{ATE}]{\tableSNFiftyThreeHundredMA};
\addplot[domain=0:100, VGGTMarker] {0.175}; %
\addplot[domain=0:100, PIMarker]  {0.150}; %
\addplot[domain=0:100, MAMarker] {0.244}; %
\addplot[domain=0:100, FasterMarker] {0.817}; %
\addplot[domain=0:100, CuterMarker]  {0.655}; %
\addplot[domain=0:100, FlareMarker] {0.419}; %

\nextgroupplot[ylabel={}, xlabel={}]
\node[anchor=north west] at (rel axis cs:0,1) {500 Frames};
\addplot[SVGGTMarker]  table[x=sparsity, y expr=\thisrow{ATE}]{\tableSNFiftyFiveHundredVGGT};
\addplot[SPIMarker]   table[x=sparsity, y expr=\thisrow{ATE}]{\tableSNFiftyFiveHundredPI};
\addplot[SMAMarker]   table[x=sparsity, y expr=\thisrow{ATE}]{\tableSNFiftyFiveHundredMA};
\addplot[domain=0:100, VGGTMarker] {0.178}; %
\addplot[domain=0:100, PIMarker]  {0.150}; %
\addplot[domain=0:100, MAMarker] {0.261}; %
\addplot[domain=0:100, FasterMarker] {0.878}; %
\addplot[domain=0:100, CuterMarker]  {0.831}; %
\addplot[domain=0:100, FlareMarker] {0.426}; %

\nextgroupplot[ylabel={}, xlabel={}]
\node[anchor=north west] at (rel axis cs:0,1) {1000 Frames};
\addplot[SVGGTMarker]  table[x=sparsity, y expr=\thisrow{ATE}]{\tableSNFiftyOneThousandVGGT};
\addplot[SPIMarker]   table[x=sparsity, y expr=\thisrow{ATE}]{\tableSNFiftyOneThousandPI};
\addplot[SMAMarker]   table[x=sparsity, y expr=\thisrow{ATE}]{\tableSNFiftyOneThousandMA};
\addplot[domain=0:100, VGGTMarker] {0.210}; %
\addplot[domain=0:100, PIMarker]  {0.150}; %
\addplot[domain=0:100, MAMarker] {0.287}; %

\nextgroupplot[ylabel={Speedup $\uparrow$}, ymin=0, ymax=3.9]
\node[anchor=north west] at (rel axis cs:0,1) {100 Frames};
\addplot[SVGGTMarker]  table[x=sparsity, y expr=4.9 / \thisrow{time}]{\tableSNFiftyOneHundredVGGT};
\addplot[SPIMarker]   table[x=sparsity, y expr=3.9 / \thisrow{time}]{\tableSNFiftyOneHundredPI};
\addplot[SMAMarker]   table[x=sparsity, y expr=2.2 / (\thisrow{time} - 19.0)]{\tableSNFiftyOneHundredMA};

\nextgroupplot[ylabel={}, ymin=0, ymax=3.9]
\node[anchor=north west] at (rel axis cs:0,1) {300 Frames};
\addplot[SVGGTMarker]  table[x=sparsity, y expr=35.1 / \thisrow{time}]{\tableSNFiftyThreeHundredVGGT};
\addplot[SPIMarker]   table[x=sparsity, y expr=27.0 / \thisrow{time}]{\tableSNFiftyThreeHundredPI};
\addplot[SMAMarker]   table[x=sparsity, y expr=14.1 / (\thisrow{time} - 55)]{\tableSNFiftyThreeHundredMA};

\nextgroupplot[ylabel={}, ymin=0, ymax=3.9]
\node[anchor=north west] at (rel axis cs:0,1) {500 Frames};
\addplot[SVGGTMarker]  table[x=sparsity, y expr=92.9 / \thisrow{time}]{\tableSNFiftyFiveHundredVGGT};
\addplot[SPIMarker]   table[x=sparsity, y expr=71.1 / \thisrow{time}]{\tableSNFiftyFiveHundredPI};
\addplot[SMAMarker]   table[x=sparsity, y expr=36.3 / (\thisrow{time} - 90.0)]{\tableSNFiftyFiveHundredMA};

\nextgroupplot[ylabel={}, ymin=0, ymax=3.9]
\node[anchor=north west] at (rel axis cs:0,1) {1000 Frames};
\addplot[SVGGTMarker]  table[x=sparsity, y expr=338.0 / \thisrow{time}]{\tableSNFiftyOneThousandVGGT};
\addlegendentry{Sparse VGGT}
\addplot[SPIMarker]   table[x=sparsity, y expr=255.6 / \thisrow{time}]{\tableSNFiftyOneThousandPI};
\addlegendentry{Sparse $\pi^3$}
\addplot[SMAMarker]   table[x=sparsity, y expr=127.6 / (\thisrow{time} - 190)]{\tableSNFiftyOneThousandMA};
\addlegendentry{Sparse MapAnything}

\coordinate (c2) at (rel axis cs:1,1);
\end{groupplot}

\coordinate (c3) at ($(c1)!.5!(c2)$);
    \node[below] at (c3 |- current bounding box.south) {\pgfplotslegendfromname{SNLongLegend}};
\end{tikzpicture}
    \caption{
        \textbf{Pose estimation and speed-up for long sequences.}
        Evaluated on a 50-sequence subset of the ScanNet~\cite{dai2017scannet} validation set; frames are evenly spaced along the entire sequence.
        This benchmark is much harder than the experiments in the main text, since the trajectories are longer and more complex.
        Our method achieves a 2$\times$ speed-up at 60\% effective sparsity with little to no drop in task performance compared to the original model.
        On longer sequences, end-to-end inference time is reduced up to 3$\times$.
        Inaccurate time measurements for MapAnything are due to extensive pre- and post-processing applied by the model.
    }
    \label{fig:sup:regression-sn-long}
\end{figure*}

\section{Ablations}
We present the results for two ablations of our method.
In the first ablation, we evaluate whether it is necessary to distinguish between patch tokens and special tokens.
In the second ablation, we investigate whether training linear projections on top of the pooled queries and keys improves the robustness of the method to sparsity.

\subsection{Treating the special tokens special}
In the main text, we distinguish between the special tokens, camera embedding and register tokens, and the patch tokens.
We apply the block-sparse attention only on the patch-to-patch attention, while we compute the special-to-patch, patch-to-special, and the special-to-special attention as usual, \ie dense.
In this experiment, we compare this approach with an implementation that does not distinguish between the camera embeddings, register tokens, and patch tokens.
The results are shown in~\cref{fig:sup:ablation-special-token}.
At low sparsity levels, the chance of skipping computations concerning special tokens is small, since there are far more patch tokens than special tokens, and both methods perform similarly.
At high sparsity levels, however, our strategy of always keeping all interactions concerning the camera and register tokens significantly reduces the performance degradation compared to the naive strategy.

\begin{figure*}
    \centering
    \begin{tikzpicture}

    \pgfplotstableread[col sep=comma]{fig/ablation_special_token/relpose_scannet.csv}\tableScanNetNoSpecial
    \pgfplotstableread[col sep=comma]{fig/rawdata/scannet/sparse_scannet_vggt.csv}\tableScanNetVGGT
    
    \pgfplotstableread[col sep=comma]{fig/ablation_special_token/relpose_co3d.csv}\tableCOThreeDNoSpecial
    \pgfplotstableread[col sep=comma]{fig/rawdata/co3d/sparse_co3d_vggt.csv}\tableCOThreeDVGGT
    
    \pgfplotstableread[col sep=comma]{fig/ablation_special_token/mv_recon_eth3d.csv}\tableETHNoSpecial
    \pgfplotstableread[col sep=comma]{fig/rawdata/eth3d/sparse_eth3d_vggt.csv}\tableETHVGGT

    \pgfplotstableread[col sep=comma]{fig/ablation_special_token/mv_recon_nrgbd.csv}\tableNRGBDNoSpecial
    \pgfplotstableread[col sep=comma]{fig/rawdata/nrgbd/sparse_nrgbd_vggt.csv}\tableNRGBDVGGT
    
    \begin{groupplot}[
        group style={
            group size=2 by 2,
            horizontal sep=1.5cm,
            vertical sep=1.5cm,
        },
        width=8cm,
        height=4cm,
        axis lines=left,
        xlabel={Avg sparsity},
        grid=both,
        grid style={gray!20},
        xmin=0, xmax=90,
        scaled y ticks=false,
        tick label style={/pgf/number format/fixed},
        legend style={
            draw=none,
            at={(-0.5, -0.5)},
            anchor=north,
            legend columns=5,
            font=\small,
            /tikz/every even column/.append style={column sep=3pt}
        },
        legend to name=regressionMVReconLegend,
    ]
    
    \nextgroupplot[title={ScanNet}, ymin=0.02, ymax=0.07, xlabel={}, ylabel={ATE}]
    \coordinate (c1) at (rel axis cs:0,1);
    \addplot[only marks, thick, mark=x, blue]  table[x=sparsity, y={ATE}]{\tableScanNetVGGT};
    \addplot[only marks, thick, mark=x, orange]  table[x=sparsity, y={ATE}]{\tableScanNetNoSpecial};

    \nextgroupplot[title={CO3Dv2}, ylabel={}, ymin=75, ymax=95, xlabel={}, ylabel={AUC@30}]
    \addplot[only marks, thick, mark=x, blue]  table[x=sparsity, y expr=100*\thisrow{AUC30}]{\tableCOThreeDVGGT};
    \addplot[only marks, thick, mark=x, orange]  table[x=sparsity, y expr=100*\thisrow{AUC30}]{\tableCOThreeDNoSpecial};

    \nextgroupplot[title={ETH3D}, ylabel={}, ymin=0.1, ymax=0.5, ylabel={Chamfer Dist.}]
    \addplot[only marks, thick, mark=x, blue]  table[x=sparsity, y=cham]{\tableETHVGGT};
    \addplot[only marks, thick, mark=x, orange]  table[x=sparsity, y=cham]{\tableETHNoSpecial};

    \nextgroupplot[title={NRGBD}, ylabel={}, ymin=0.02, ymax=0.15, ylabel={Chamfer Dist.}]
    \addplot[only marks, thick, mark=x, blue]  table[x=sparsity, y=cham]{\tableNRGBDVGGT};
    \addlegendentry{Separate Attention for Special Tokens}
    \addplot[only marks, thick, mark=x, orange]  table[x=sparsity, y=cham]{\tableNRGBDNoSpecial};
    \addlegendentry{Block-Sparse over all tokens}
    
    \coordinate (c2) at (rel axis cs:1,1);
    \end{groupplot}
    
    \coordinate (c3) at ($(c1)!.5!(c2)$);
        \node[below] at (c3 |- current bounding box.south) {\pgfplotslegendfromname{regressionMVReconLegend}};
    \end{tikzpicture}
    \caption{
        \textbf{Ablation on special treatment of special tokens.}
        We compare the downstream performance of simple block-sparse attention over all tokens against our approach of separating patch tokens from camera and register tokens.
        For higher sparsity ratios, the chance of dropping special tokens rises, and at the same time, the task performance drops significantly more sharply for the simplified variant.
    }
    \label{fig:sup:ablation-special-token}
\end{figure*}

\subsection{Learning additional linear projections}
We train additional linear projection layers on top of the pooled key and query representations following SeerAttention~\cite{gao2024seerattention}.
At training time, the linear projections are optimized to predict the entries of the downsampled attention map; at inference time, the predicted low-res attention map is converted into a binary block mask, as in the main text.
With this experiment, we investigate whether the linear projections improve the robustness of the model for higher sparsity ratios compared to the default key and query representations.

Our training setup follows SeerAttention~\cite{gao2024seerattention}, that is, one linear projection per attention head.
The training objective is a self-supervised KL loss between the downsampled ground-truth attention matrix and the logits of the predicted low-res mask.
We train on BlendedMVS~\cite{yao2020blendedmvs} because it contains real indoor and outdoor scenes and is similar to VGGT's training data.
Since no gradients need to be passed through the actual model, training is both fast and lightweight.
We train the projection layers for 3k steps with a batch size of 16, and a sequence length of 8 frames at resolutions sampled from $518^2$ and $518\times 378$, using the AdamW optimizer with a learning rate of $10^{-3}$ and weight decay of $0.01$.
The block size is set to 128 for the queries and 64 for the keys.
Training finishes in around three hours on a single H100.

The results shown in~\cref{fig:sup:ablation-learned-projection} show little to no improvements over the training-free baseline.

\begin{figure*}[p]
    \centering
    \begin{tikzpicture}

    \pgfplotstableread[col sep=comma]{fig/ablation_learned_projection/relpose_scannet.csv}\tableScanNetTrained
    \pgfplotstableread[col sep=comma]{fig/rawdata/scannet/sparse_scannet_vggt.csv}\tableScanNetVGGT
    
    \pgfplotstableread[col sep=comma]{fig/ablation_learned_projection/relpose_co3d.csv}\tableCOThreeDTrained
    \pgfplotstableread[col sep=comma]{fig/rawdata/co3d/sparse_co3d_vggt.csv}\tableCOThreeDVGGT
    
    \pgfplotstableread[col sep=comma]{fig/ablation_learned_projection/mv_recon_eth3d.csv}\tableETHTrained
    \pgfplotstableread[col sep=comma]{fig/rawdata/eth3d/sparse_eth3d_vggt.csv}\tableETHVGGT

    \pgfplotstableread[col sep=comma]{fig/ablation_learned_projection/mv_recon_nrgbd.csv}\tableNRGBDTrained
    \pgfplotstableread[col sep=comma]{fig/rawdata/nrgbd/sparse_nrgbd_vggt.csv}\tableNRGBDVGGT
    
    \begin{groupplot}[
        group style={
            group size=2 by 2,
            horizontal sep=1.5cm,
            vertical sep=1.5cm,
        },
        width=8cm,
        height=4cm,
        axis lines=left,
        xlabel={Avg sparsity},
        grid=both,
        grid style={gray!20},
        xmin=0, xmax=90,
        scaled y ticks=false,
        tick label style={/pgf/number format/fixed},
        legend style={
            draw=none,
            at={(-0.5, -0.5)},
            anchor=north,
            legend columns=5,
            font=\small,
            /tikz/every even column/.append style={column sep=3pt}
        },
        legend to name=regressionMVReconLegend,
    ]
    
    \nextgroupplot[title={ScanNet}, ymin=0.02, ymax=0.07, xlabel={}, ylabel={ATE}]
    \coordinate (c1) at (rel axis cs:0,1);
    \addplot[only marks, thick, mark=x, blue]  table[x=sparsity, y={ATE}]{\tableScanNetVGGT};
    \addplot[only marks, thick, mark=x, orange]  table[x=sparsity, y={ATE}]{\tableScanNetTrained};

    \nextgroupplot[title={CO3Dv2}, ylabel={}, ymin=75, ymax=95, xlabel={}, ylabel={AUC@30}]
    \addplot[only marks, thick, mark=x, blue]  table[x=sparsity, y expr=100*\thisrow{AUC30}]{\tableCOThreeDVGGT};
    \addplot[only marks, thick, mark=x, orange]  table[x=sparsity, y expr=100*\thisrow{AUC30}]{\tableCOThreeDTrained};

    \nextgroupplot[title={ETH3D}, ylabel={}, ymin=0.1, ymax=0.5, ylabel={Chamfer Dist.}]
    \addplot[only marks, thick, mark=x, blue]  table[x=sparsity, y=cham]{\tableETHVGGT};
    \addplot[only marks, thick, mark=x, orange]  table[x=sparsity, y=cham]{\tableETHTrained};

    \nextgroupplot[title={NRGBD}, ylabel={}, ymin=0.02, ymax=0.15, ylabel={Chamfer Dist.}]
    \addplot[only marks, thick, mark=x, blue]  table[x=sparsity, y=cham]{\tableNRGBDVGGT};
    \addlegendentry{Training-free}
    \addplot[only marks, thick, mark=x, orange]  table[x=sparsity, y=cham]{\tableNRGBDTrained};
    \addlegendentry{Trained linear projections}
    
    \coordinate (c2) at (rel axis cs:1,1);
    \end{groupplot}
    
    \coordinate (c3) at ($(c1)!.5!(c2)$);
        \node[below] at (c3 |- current bounding box.south) {\pgfplotslegendfromname{regressionMVReconLegend}};
    \end{tikzpicture}
    \caption{
        \textbf{Ablation on learned linear projections.}
        We compare the downstream performance of training-free block-sparse mask prediction with an approach following SeerAttention~\cite{gao2024seerattention}, that is, training linear projections on top of pooled query and key features.
        The model with additional trained linear layers does not show significantly improved performance or robustness against increased sparsity compared to the training-free variant.
    }
    \label{fig:sup:ablation-learned-projection}
\end{figure*}

\subsection{Full Layer Skip Ablations}
We further investigate the impact of skipping global attention layers in the style of~\cref{fig:layer-drop} in the main text.
Results for VGGT are shown in~\cref{fig:sup:layer-drop-vggt}, and for MapAnything in~\cref{fig:sup:layer-drop-ma}.
We do omit $\pi^3$ because it behaves very similar to VGGT.

For VGGT, the experiments indicate a strong sensitivity of the model to perturbations in the middle layers, regardless of the task.
On Co3Dv2, for example, removing just four (out of 24) global attention layers in the middle of the aggregator reduces VGGT's task performance from state-of-the-art levels to zero AUC@30.
Removing four of the first or last global attention layers, in contrast, retains most of the task performance on ScanNet, ETH3D, and NRGBD.
On Co3Dv2, we observe a moderate performance drop when removing layers towards the model output, but a much smaller drop when removing layers early in the aggregator.

In contrast to VGGT, MapAnything shows much more robust performance when we remove layers in the middle of the aggregator, and task performance seems to rely much more on the last few global attention layers.
Removing the first nine (out of twelve) global attention layers retains the same task performance as removing the last layer.
We hypothesize that this behaviour stems from the MapAnything's architecture, which uses the image patch tokens to predict camera poses instead of special camera tokens like VGGT.

For both models, the performance drops due to layer skipping are much more severe than the drop that we observe when applying our block-sparse attention scheme.

\begin{figure*}
\centering
\begin{tikzpicture}

\pgfplotstableread[col sep=comma]{fig/layer_drop/layer_drop_eth3d_both.csv}\tableSkipBothETH
\pgfplotstableread[col sep=comma]{fig/layer_drop/layer_drop_eth3d_front.csv}\tableSkipFrontETH
\pgfplotstableread[col sep=comma]{fig/layer_drop/layer_drop_eth3d_middle.csv}\tableSkipMiddleETH
\pgfplotstableread[col sep=comma]{fig/layer_drop/layer_drop_eth3d_back.csv}\tableSkipBackETH

\pgfplotstableread[col sep=comma]{fig/layer_drop/layer_drop_co3d_both.csv}\tableSkipBothCOThreeD
\pgfplotstableread[col sep=comma]{fig/layer_drop/layer_drop_co3d_front.csv}\tableSkipFrontCOThreeD
\pgfplotstableread[col sep=comma]{fig/layer_drop/layer_drop_co3d_middle.csv}\tableSkipMiddleCOThreeD
\pgfplotstableread[col sep=comma]{fig/layer_drop/layer_drop_co3d_back.csv}\tableSkipBackCOThreeD

\pgfplotstableread[col sep=comma]{fig/layer_drop/layer_drop_NRGBD_both.csv}\tableSkipBothNRGBD
\pgfplotstableread[col sep=comma]{fig/layer_drop/layer_drop_NRGBD_front.csv}\tableSkipFrontNRGBD
\pgfplotstableread[col sep=comma]{fig/layer_drop/layer_drop_NRGBD_middle.csv}\tableSkipMiddleNRGBD
\pgfplotstableread[col sep=comma]{fig/layer_drop/layer_drop_NRGBD_back.csv}\tableSkipBackNRGBD

\pgfplotstableread[col sep=comma]{fig/layer_drop/layer_drop_scannet_both.csv}\tableSkipBothScanNet
\pgfplotstableread[col sep=comma]{fig/layer_drop/layer_drop_scannet_front.csv}\tableSkipFrontScanNet
\pgfplotstableread[col sep=comma]{fig/layer_drop/layer_drop_scannet_middle.csv}\tableSkipMiddleScanNet
\pgfplotstableread[col sep=comma]{fig/layer_drop/layer_drop_scannet_back.csv}\tableSkipBackScanNet

\begin{groupplot}[
    group style={
        group size=2 by 2,
        horizontal sep=1.5cm,
        vertical sep=1.5cm,
    },
    width=8cm,
    height=4cm,
    xmin=0, xmax=24.5,
    axis lines=left,
    xlabel={\#Skipped global attention},
    tick label style={font=\small},
    label style={font=\small},
    legend style={
        draw=none,
        font=\small,
        at={(-0.5, -0.5)},
        anchor=north,
        legend columns=-1,
        row sep=1pt,
        /tikz/every even column/.append style={column sep=3pt}
    },
    legend to name=layerDropAllLegend,
    legend cell align=left,
    clip=false,
]

\nextgroupplot[xlabel={}, ymin=0, ymax=0.5, ylabel={ATE}, title={ScanNet}]
\coordinate (c1) at (rel axis cs:0,1);

\addplot[
    dashed,
    black,
    domain=0:24
] {0.035};

\addplot[mark=*, line width=1.1pt, color=Paired-B] table[x=numskipped, y={ATE}]{\tableSkipFrontScanNet};
\addplot[mark=*, line width=1.1pt, color=Paired-D] table[x=numskipped,y={ATE}]{\tableSkipBackScanNet};
\addplot[mark=*, line width=1.1pt, color=Paired-F] table[x=numskipped,y={ATE}]{\tableSkipBothScanNet};
\addplot[mark=*, line width=1.1pt, color=Paired-H] table[x=numskipped,y={ATE}]{\tableSkipMiddleScanNet};

\nextgroupplot[xlabel={}, ymin=0, ymax=1, ylabel={AUC@30}, title={CO3Dv2}]

\addplot[
    dashed,
    black,
    domain=0:24
] {0.909};

\addplot[mark=*, line width=1.1pt, color=Paired-B] table[x=numskipped, y={AUC30}]{\tableSkipFrontCOThreeD};
\addplot[mark=*, line width=1.1pt, color=Paired-D] table[x=numskipped,y={AUC30}]{\tableSkipBackCOThreeD};
\addplot[mark=*, line width=1.1pt, color=Paired-F] table[x=numskipped,y={AUC30}]{\tableSkipBothCOThreeD};
\addplot[mark=*, line width=1.1pt, color=Paired-H] table[x=numskipped,y={AUC30}]{\tableSkipMiddleCOThreeD};

\nextgroupplot[ymin=0, ymax=2, ylabel={Chamfer Dist.}, title={ETH3D}]

\addplot[
    dashed,
    black,
    domain=0:24
] {0.203};

\addplot[mark=*, line width=1.1pt, color=Paired-B] table[x=numskipped, y={cham}]{\tableSkipFrontETH};
\addplot[mark=*, line width=1.1pt, color=Paired-D] table[x=numskipped,y={cham}]{\tableSkipBackETH};
\addplot[mark=*, line width=1.1pt, color=Paired-F] table[x=numskipped,y={cham}]{\tableSkipBothETH};
\addplot[mark=*, line width=1.1pt, color=Paired-H] table[x=numskipped,y={cham}]{\tableSkipMiddleETH};

\nextgroupplot[ymin=0, ymax=1, ylabel={Chamfer Dist.}, title={NRGBD}]

\addplot[
    dashed,
    black,
    domain=0:24
] {0.048};
\addlegendentry{None}

\addplot[mark=*, line width=1.1pt, color=Paired-B] table[x=numskipped, y={cham}]{\tableSkipFrontNRGBD};
\addlegendentry{Front}

\addplot[mark=*, line width=1.1pt, color=Paired-D] table[x=numskipped,y={cham}]{\tableSkipBackNRGBD};
\addlegendentry{Back}

\addplot[mark=*, line width=1.1pt, color=Paired-F] table[x=numskipped,y={cham}]{\tableSkipBothNRGBD};
\addlegendentry{Front \& Back}

\addplot[mark=*, line width=1.1pt, color=Paired-H] table[x=numskipped,y={cham}]{\tableSkipMiddleNRGBD};
\addlegendentry{Mid}

\coordinate (c2) at (rel axis cs:1,1);
\end{groupplot}

\coordinate (c3) at ($(c1)!.5!(c2)$);
    \node[below] at (c3 |- current bounding box.south) {\pgfplotslegendfromname{layerDropAllLegend}};
\end{tikzpicture}
\caption{
    \textbf{VGGT is sensitive to pruning of the middle aggregator layers.}
    We skip the computation of different global attention layers in the aggregator starting with the earliest (Front), last (Back), alternating (Front \& Back), or from the middle layers (Middle), and evaluate the performance drop on different tasks.
    The x-axis denotes the total number of skipped layers.
    The experiment shows that VGGT is especially sensitive to pruning of the center layers, and robust against pruning the early and late layers.
    MapAnything, in contrast, is much more sensitive to alterations in the last layers.
}
\label{fig:sup:layer-drop-vggt}
\end{figure*}

\begin{figure*}
\centering
\begin{tikzpicture}

\pgfplotstableread[col sep=comma]{fig/layer_drop/layer_drop_eth3d_both_ma.csv}\tableSkipBothETH
\pgfplotstableread[col sep=comma]{fig/layer_drop/layer_drop_eth3d_front_ma.csv}\tableSkipFrontETH
\pgfplotstableread[col sep=comma]{fig/layer_drop/layer_drop_eth3d_middle_ma.csv}\tableSkipMiddleETH
\pgfplotstableread[col sep=comma]{fig/layer_drop/layer_drop_eth3d_back_ma.csv}\tableSkipBackETH

\pgfplotstableread[col sep=comma]{fig/layer_drop/layer_drop_co3d_both_ma.csv}\tableSkipBothCOThreeD
\pgfplotstableread[col sep=comma]{fig/layer_drop/layer_drop_co3d_front_ma.csv}\tableSkipFrontCOThreeD
\pgfplotstableread[col sep=comma]{fig/layer_drop/layer_drop_co3d_middle_ma.csv}\tableSkipMiddleCOThreeD
\pgfplotstableread[col sep=comma]{fig/layer_drop/layer_drop_co3d_back_ma.csv}\tableSkipBackCOThreeD

\pgfplotstableread[col sep=comma]{fig/layer_drop/layer_drop_NRGBD_both_ma.csv}\tableSkipBothNRGBD
\pgfplotstableread[col sep=comma]{fig/layer_drop/layer_drop_NRGBD_front_ma.csv}\tableSkipFrontNRGBD
\pgfplotstableread[col sep=comma]{fig/layer_drop/layer_drop_NRGBD_middle_ma.csv}\tableSkipMiddleNRGBD
\pgfplotstableread[col sep=comma]{fig/layer_drop/layer_drop_NRGBD_back_ma.csv}\tableSkipBackNRGBD

\pgfplotstableread[col sep=comma]{fig/layer_drop/layer_drop_scannet_both_ma.csv}\tableSkipBothScanNet
\pgfplotstableread[col sep=comma]{fig/layer_drop/layer_drop_scannet_front_ma.csv}\tableSkipFrontScanNet
\pgfplotstableread[col sep=comma]{fig/layer_drop/layer_drop_scannet_middle_ma.csv}\tableSkipMiddleScanNet
\pgfplotstableread[col sep=comma]{fig/layer_drop/layer_drop_scannet_back_ma.csv}\tableSkipBackScanNet

\begin{groupplot}[
    group style={
        group size=2 by 2,
        horizontal sep=1.5cm,
        vertical sep=1.5cm,
    },
    width=8cm,
    height=4cm,
    xmin=0, xmax=12.5,
    axis lines=left,
    xlabel={\#Skipped global attention},
    tick label style={font=\small},
    label style={font=\small},
    legend style={
        draw=none,
        font=\small,
        at={(-0.5, -0.5)},
        anchor=north,
        legend columns=-1,
        row sep=1pt,
        /tikz/every even column/.append style={column sep=3pt}
    },
    legend to name=layerDropAllLegend,
    legend cell align=left,
    clip=false,
]

\nextgroupplot[xlabel={}, ymin=0, ymax=0.5, ylabel={ATE}, title={ScanNet}]
\coordinate (c1) at (rel axis cs:0,1);

\addplot[
    dashed,
    black,
    domain=0:12
] {0.061};

\addplot[mark=*, line width=1.1pt, color=Paired-B] table[x=numskipped, y={ATE}]{\tableSkipFrontScanNet};
\addplot[mark=*, line width=1.1pt, color=Paired-D] table[x=numskipped,y={ATE}]{\tableSkipBackScanNet};
\addplot[mark=*, line width=1.1pt, color=Paired-F] table[x=numskipped,y={ATE}]{\tableSkipBothScanNet};
\addplot[mark=*, line width=1.1pt, color=Paired-H] table[x=numskipped,y={ATE}]{\tableSkipMiddleScanNet};

\nextgroupplot[xlabel={}, ymin=0, ymax=1, ylabel={AUC@30}, title={CO3Dv2}]

\addplot[
    dashed,
    black,
    domain=0:12
] {0.694};

\addplot[mark=*, line width=1.1pt, color=Paired-B] table[x=numskipped, y={AUC30}]{\tableSkipFrontCOThreeD};
\addplot[mark=*, line width=1.1pt, color=Paired-D] table[x=numskipped,y={AUC30}]{\tableSkipBackCOThreeD};
\addplot[mark=*, line width=1.1pt, color=Paired-F] table[x=numskipped,y={AUC30}]{\tableSkipBothCOThreeD};
\addplot[mark=*, line width=1.1pt, color=Paired-H] table[x=numskipped,y={AUC30}]{\tableSkipMiddleCOThreeD};

\nextgroupplot[ymin=0, ymax=1.3, ylabel={Chamfer Dist.}, title={ETH3D}]

\addplot[
    dashed,
    black,
    domain=0:12
] {0.194};

\addplot[mark=*, line width=1.1pt, color=Paired-B] table[x=numskipped, y={cham}]{\tableSkipFrontETH};
\addplot[mark=*, line width=1.1pt, color=Paired-D] table[x=numskipped,y={cham}]{\tableSkipBackETH};
\addplot[mark=*, line width=1.1pt, color=Paired-F] table[x=numskipped,y={cham}]{\tableSkipBothETH};
\addplot[mark=*, line width=1.1pt, color=Paired-H] table[x=numskipped,y={cham}]{\tableSkipMiddleETH};

\nextgroupplot[ymin=0, ymax=0.5, ylabel={Chamfer Dist.}, title={NRGBD}]

\addplot[
    dashed,
    black,
    domain=0:12
] {0.093};
\addlegendentry{None}

\addplot[mark=*, line width=1.1pt, color=Paired-B] table[x=numskipped, y={cham}]{\tableSkipFrontNRGBD};
\addlegendentry{Front}

\addplot[mark=*, line width=1.1pt, color=Paired-D] table[x=numskipped,y={cham}]{\tableSkipBackNRGBD};
\addlegendentry{Back}

\addplot[mark=*, line width=1.1pt, color=Paired-F] table[x=numskipped,y={cham}]{\tableSkipBothNRGBD};
\addlegendentry{Front \& Back}

\addplot[mark=*, line width=1.1pt, color=Paired-H] table[x=numskipped,y={cham}]{\tableSkipMiddleNRGBD};
\addlegendentry{Mid}

\coordinate (c2) at (rel axis cs:1,1);
\end{groupplot}

\coordinate (c3) at ($(c1)!.5!(c2)$);
    \node[below] at (c3 |- current bounding box.south) {\pgfplotslegendfromname{layerDropAllLegend}};
\end{tikzpicture}
\caption{
    \textbf{MapAnything is sensitive to pruning of the later aggregator layers.}
    We skip the computation of different global attention layers in the aggregator starting with the earliest (Front), last (Back), alternating (Front \& Back), or from the middle layers (Middle), and evaluate the performance drop on different tasks.
    The x-axis denotes the total number of skipped layers.
    MapAnything, in contrast to VGGT, is sensitive to alterations in the last aggregator layers.
}
\label{fig:sup:layer-drop-ma}
\end{figure*}

\section{Additional Visualizations}

\subsection{Additional Qualitative Results}
We provide further visualizations of reconstructed point clouds and estimated trajectories for all three tested models in~\cref{fig:sup:qualitative}.

\begin{figure*}
    \centering
    \input{fig/qualitative_samples_sup.tex}
    \caption{
        \textbf{Qualitative point map estimation results.}
        We show the reconstructed pointcloud of the \textit{terrace} scene of the ETH3D~\cite{schoeps2017eth3d} dataset.
        Increasing sparsity leads to small perturbations in the reconstructed pointclouds, but the overall structure of the reconstructed scene stays consistent.
    }
    \label{fig:sup:qualitative}
\end{figure*}

\begin{figure*}
    \centering
    \input{fig/qualitative_samples_pose_sup.tex}
    \caption{
        \textbf{Qualitative long-sequence pose estimation results.}
        We show the estimated camera trajectory of a 1000-frame sequence of the ScanNet~\cite{dai2017scannet} validation set.
        Even with high sparsity, the estimated camera poses are similar to the poses estimated by the original model.
    }
    \label{fig:sup:qualitative-pose}
\end{figure*}

\subsection{Attention map statistics}
We plot the average and maximum value of the global attention matrix for layers in the aggregator of VGGT, $\pi^3$, and MapAnything in~\cref{fig:sup:attention-distribution-all}, in the same style as~\cref{fig:sparse-global-attention} in the main text.
While $\pi^3$ uses less global attention layers in the aggregator than VGGT, the statistics of the remaining layers closely resemble those of the corresponding layers in VGGT.
We hypothesize that the reason for this remarkable similarity is the fact that $\pi^3$ was finetuned from a VGGT checkpoint~\cite{wang2025pi}.
MapAnything shows consistently high maximum activations in the patch-to-patch attention in all layers.
Compared to VGGT and $\pi^3$, the peak maximum activation is lower.
The average special-to-special attention score is much higher than for the other models, and we hypothesize that this is due to the fact that MapAnything only employs a single special token (the metric scale token).

\begin{figure*}
    \centering
    \begin{tikzpicture}
\pgfplotstableread[col sep=comma]{fig/attn_stats/attn_stats_vggt.csv}\tableAttnStatsVGGT
\pgfplotstableread[col sep=comma]{fig/attn_stats/attn_stats_pi3.csv}\tableAttnStatsPI
\pgfplotstableread[col sep=comma]{fig/attn_stats/attn_stats_ma.csv}\tableAttnStatsMA

\begin{groupplot}[
    group style={
        group size=3 by 2,
        vertical sep=4pt,
        horizontal sep=4pt,
        yticklabels at=edge left,
    },
    width=6.8cm, height=3.9cm,
    ymode=log,
    xmin=0.1, xmax=12.9,
    xlabel={Global Attention Layer},
    ylabel={Activation},
    legend style={
        draw=none,
        font=\small,
        at={(0.5,-0.55)},
        anchor=north,
        legend columns=4,
        /tikz/every even column/.append style={column sep=3pt},
    },
    legend to name=attentionStatsLegend,
]

\nextgroupplot[xticklabels=\empty, xlabel={}, ylabel={Max Entry}, ymin=0.0002, ymax=0.75, title={VGGT}, xmax=24.9]
\coordinate (c1) at (rel axis cs:0,1);

\adderrorzone[x expr=\coordindex+1, y name=s2sMaxMean, y err name=s2sMaxStd] {\tableAttnStatsVGGT}{Paired-B}{Paired-A};
\adderrorzone[x expr=\coordindex+1, y name=s2pMaxMean, y err name=s2pMaxStd] {\tableAttnStatsVGGT}{Paired-D}{Paired-C};
\adderrorzone[x expr=\coordindex+1, y name=p2pMaxMean, y err name=p2pMaxStd] {\tableAttnStatsVGGT}{Paired-F}{Paired-E};
\adderrorzone[x expr=\coordindex+1, y name=p2sMaxMean, y err name=p2sMaxStd] {\tableAttnStatsVGGT}{Paired-H}{Paired-G};

\nextgroupplot[xticklabels=\empty, xlabel={}, ylabel={}, ymin=0.0002, ymax=0.75, title={$\pi^3$}, xmax=18.9]

\adderrorzone[x expr=\coordindex+1, y name=s2sMaxMean, y err name=s2sMaxStd] {\tableAttnStatsPI}{Paired-B}{Paired-A};
\adderrorzone[x expr=\coordindex+1, y name=s2pMaxMean, y err name=s2pMaxStd] {\tableAttnStatsPI}{Paired-D}{Paired-C};
\adderrorzone[x expr=\coordindex+1, y name=p2pMaxMean, y err name=p2pMaxStd] {\tableAttnStatsPI}{Paired-F}{Paired-E};
\adderrorzone[x expr=\coordindex+1, y name=p2sMaxMean, y err name=p2sMaxStd] {\tableAttnStatsPI}{Paired-H}{Paired-G};

\nextgroupplot[xticklabels=\empty, xlabel={}, ylabel={}, ymin=0.0002, ymax=0.75, title={MapAnything}]

\adderrorzone[x expr=\coordindex+1, y name=s2sMaxMean, y err name=s2sMaxStd] {\tableAttnStatsMA}{Paired-B}{Paired-A};
\adderrorzone[x expr=\coordindex+1, y name=s2pMaxMean, y err name=s2pMaxStd] {\tableAttnStatsMA}{Paired-D}{Paired-C};
\adderrorzone[x expr=\coordindex+1, y name=p2pMaxMean, y err name=p2pMaxStd] {\tableAttnStatsMA}{Paired-F}{Paired-E};
\adderrorzone[x expr=\coordindex+1, y name=p2sMaxMean, y err name=p2sMaxStd] {\tableAttnStatsMA}{Paired-H}{Paired-G};

\nextgroupplot[ylabel={Avg. Entry}, xmax=24.9, ymode=log, ymin=0.00005, ymax=0.15]

\adderrorzone[x expr=\coordindex+1, y name=s2sMeanMean, y err name=s2sMeanStd] {\tableAttnStatsVGGT}{Paired-B}{Paired-A};
\adderrorzone[x expr=\coordindex+1, y name=s2pMeanMean, y err name=s2pMeanStd] {\tableAttnStatsVGGT}{Paired-D}{Paired-C};
\adderrorzone[x expr=\coordindex+1, y name=p2pMeanMean, y err name=p2pMeanStd] {\tableAttnStatsVGGT}{Paired-F}{Paired-E};
\adderrorzone[x expr=\coordindex+1, y name=p2sMeanMean, y err name=p2sMeanStd] {\tableAttnStatsVGGT}{Paired-H}{Paired-G};

\nextgroupplot[ylabel={}, xmax=18.9, ymode=log, ymin=0.00005, ymax=0.15]

\adderrorzone[x expr=\coordindex+1, y name=s2sMeanMean, y err name=s2sMeanStd] {\tableAttnStatsPI}{Paired-B}{Paired-A};
\adderrorzone[x expr=\coordindex+1, y name=s2pMeanMean, y err name=s2pMeanStd] {\tableAttnStatsPI}{Paired-D}{Paired-C};
\adderrorzone[x expr=\coordindex+1, y name=p2pMeanMean, y err name=p2pMeanStd] {\tableAttnStatsPI}{Paired-F}{Paired-E};
\adderrorzone[x expr=\coordindex+1, y name=p2sMeanMean, y err name=p2sMeanStd] {\tableAttnStatsPI}{Paired-H}{Paired-G};

\nextgroupplot[ylabel={}, ymode=log, ymin=0.00005, ymax=0.15]

\adderrorzone[x expr=\coordindex+1, y name=s2sMeanMean, y err name=s2sMeanStd] {\tableAttnStatsMA}{Paired-B}{Paired-A};
\addlegendentry{Special-to-Special}
\adderrorzone[x expr=\coordindex+1, y name=s2pMeanMean, y err name=s2pMeanStd] {\tableAttnStatsMA}{Paired-D}{Paired-C};
\addlegendentry{Special-to-Patch}
\adderrorzone[x expr=\coordindex+1, y name=p2pMeanMean, y err name=p2pMeanStd] {\tableAttnStatsMA}{Paired-F}{Paired-E};
\addlegendentry{Patch-to-Patch}
\adderrorzone[x expr=\coordindex+1, y name=p2sMeanMean, y err name=p2sMeanStd] {\tableAttnStatsMA}{Paired-H}{Paired-G};
\addlegendentry{Patch-to-Special}

\coordinate (c2) at (rel axis cs:1,1);
\end{groupplot}

\coordinate (c3) at ($(c1)!.5!(c2)$);
    \node[below=-1mm] at (c3 |- current bounding box.south) {\pgfplotslegendfromname{attentionStatsLegend}};
\end{tikzpicture}
    \caption{
        \textbf{Average \& maximum attention scores in global attention maps.}
        Note that the models use different numbers of global attention blocks in the aggregator.
        Since $\pi^3$ started training from a VGGT checkpoint, the statistics are highly similar to VGGT.
        MapAnything uses only a single special token, which explains the comparatively high average special-to-special attention score.
    }
    \label{fig:sup:attention-distribution-all}
\end{figure*}

\subsection{High-res Attention Maps}
\begin{figure*}
    \centering
    \includegraphics[width=\textwidth]{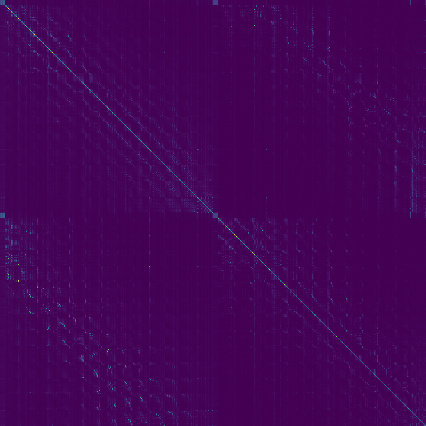}
    \caption{
    Larger visualization of the global attention matrix of aggregator layer 15 of VGGT.
    We show the average over all heads after the softmax.
    }
    \label{fig:sup:attn-maps}
\end{figure*}

We provide further visualizations of VGGT's attention maps in~\cref{fig:sup:attn-maps}.
Note that all attention map visualizations in this paper are done with inputs rescaled to at most 224px for better visibility of token activations.

\subsection{Further Correspondence Visualizations}
\begin{figure*}
    \centering
    \setlength{\tabcolsep}{2pt} %
    \centering
    \begin{tikzpicture}
        \node (vggt) at (0,0) {\includegraphics[width=0.95\linewidth]{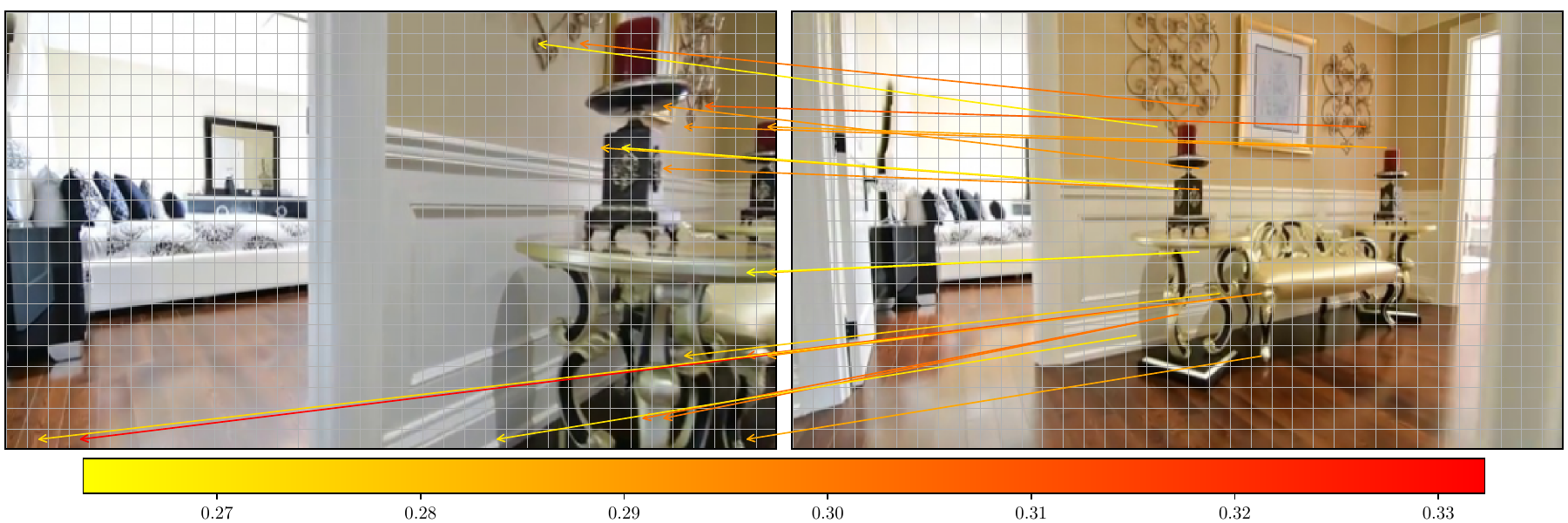}};
        
        \node[rotate=90, anchor=center] at ($(vggt.west)+(-3mm,2mm)$) {VGGT};
        
        \node[anchor=south, yshift=-2mm] at ($(vggt.north west)!0.25!(vggt.north east)$) {Query Image};
        \node[anchor=south, yshift=-2mm] at ($(vggt.north west)!0.75!(vggt.north east)$) {Key Image};
    
        \node (piii) [below=-1mm of vggt] {\includegraphics[width=0.95\linewidth]{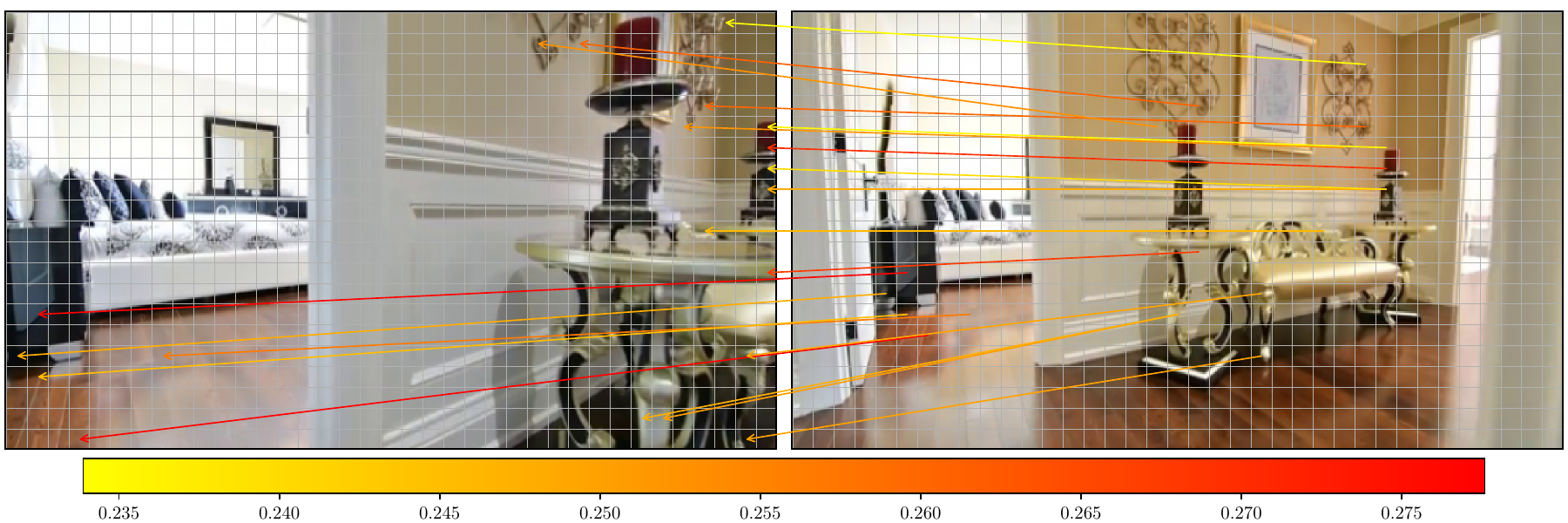}};
    
        \node[rotate=90, anchor=center] at ($(piii.west)+(-3mm,2mm)$) {$\pi^3$};
    \end{tikzpicture}

    \caption{
    Visualization of layer-15 correspondences for VGGT (top) and $\pi^3$ (bottom).
    Each row depicts a query image (left) and a key image (right), with arrows indicating matched points across views, where warmer colors indicates a higher attention value for the matches.
    Two visually identical Candles are present in both images.
    In both methods, the correspondences associated with one Candle in the key image are consistently mapped to the same Candle in the query image, rather than being confused with the second instance.
    This demonstrates the ability of the models to resolve ambiguities in scenes containing repeated objects and to maintain consist correspondences across viewpoint changes.
    }
    \label{fig:sup:correspondences}
\end{figure*}

We provide additional qualitative results for correspondence estimation in~\cref{fig:sup:correspondences}, demonstrating how VGGT and $\pi^3$ establish matches even in challenging scenarios with repeated structures and significant viewpoint changes.

\section{Full Results Tables}
For completeness, we provide the full results for all models in~\cref{fig:sparsity-regression} and~\cref{fig:regression-tandt} in the tables on the following pages.

\clearpage

\begin{table*}[p]
\centering
\begin{tabular}{lccccccc}
\toprule
Model & Acc & Comp & N.C. & $\rho$ (\%) & CDF & Sparsity \\
\midrule
VGGT & 0.050 & 0.062 & 0.750 & 10 & 0.970 & 7.630 \\
VGGT & 0.053 & 0.062 & 0.746 & 30 & 0.970 & 11.580 \\
VGGT & 0.052 & 0.061 & 0.746 & 50 & 0.970 & 13.870 \\
VGGT & 0.053 & 0.062 & 0.749 & 40 & 0.950 & 16.550 \\
VGGT & 0.057 & 0.063 & 0.747 & 60 & 0.930 & 22.310 \\
VGGT & 0.054 & 0.061 & 0.747 & 50 & 0.900 & 25.540 \\
VGGT & 0.052 & 0.059 & 0.748 & 50 & 0.880 & 28.170 \\
\bottomrule
\end{tabular}

\caption{Full results for Seven Scenes~\cite{shotton2013sevenscenes}}
\end{table*}

\begin{table*}[p]
\centering
\begin{tabular}{lccccccc}
\toprule
Model & RRA@30 & RTA@30 & AUC@30 & $\rho$ (\%) & CDF & Sparsity \\
\midrule
VGGT & 0.975 & 0.916 & 0.908 & 10 & 0.970 & 7.630 \\
VGGT & 0.975 & 0.915 & 0.908 & 30 & 0.970 & 11.580 \\
VGGT & 0.975 & 0.915 & 0.907 & 50 & 0.970 & 13.870 \\
VGGT & 0.973 & 0.914 & 0.906 & 40 & 0.950 & 16.550 \\
VGGT & 0.971 & 0.912 & 0.902 & 60 & 0.930 & 22.310 \\
VGGT & 0.967 & 0.910 & 0.899 & 50 & 0.900 & 25.540 \\
VGGT & 0.964 & 0.909 & 0.896 & 50 & 0.880 & 28.170 \\
\bottomrule
\end{tabular}

\caption{Full results for Co3Dv2~\cite{reizenstein21co3d}}
\end{table*}

\begin{table*}[p]
\centering
\begin{tabular}{lccccccc}
\toprule
Model & Acc & Comp & N.C. & $\rho$ (\%) & CDF & Sparsity \\
\midrule
VGGT & 1.075 & 1.331 & 0.613 & 10 & 0.970 & 7.774 \\
VGGT & 1.068 & 1.330 & 0.612 & 30 & 0.970 & 12.173 \\
VGGT & 1.072 & 1.335 & 0.613 & 40 & 0.950 & 17.249 \\
VGGT & 1.172 & 1.286 & 0.609 & 50 & 0.750 & 41.589 \\
VGGT & 1.215 & 1.281 & 0.609 & 50 & 0.700 & 45.062 \\
VGGT & 1.114 & 1.316 & 0.611 & 50 & 0.850 & 32.250 \\
VGGT & 1.100 & 1.328 & 0.610 & 50 & 0.880 & 28.764 \\
\bottomrule
\end{tabular}

\caption{Full results for DTU~\cite{jensen2014dtu}}
\end{table*}

\begin{table*}[p]
\centering
\begin{tabular}{lccccccc}
\toprule
Model & Acc & Comp & N.C. & $\rho$ (\%) & CDF & Sparsity \\
\midrule
VGGT & 0.232 & 0.190 & 0.886 & 10 & 0.970 & 7.630 \\
VGGT & 0.220 & 0.176 & 0.888 & 30 & 0.970 & 11.580 \\
VGGT & 0.238 & 0.180 & 0.884 & 50 & 0.970 & 13.870 \\
VGGT & 0.218 & 0.174 & 0.887 & 40 & 0.950 & 16.550 \\
VGGT & 0.231 & 0.187 & 0.874 & 60 & 0.930 & 22.310 \\
VGGT & 0.212 & 0.172 & 0.885 & 50 & 0.900 & 25.540 \\
VGGT & 0.213 & 0.169 & 0.884 & 50 & 0.880 & 28.170 \\
\bottomrule
\end{tabular}

\caption{Full results for ETH3D~\cite{schoeps2017eth3d}}
\end{table*}

\begin{table*}[p]
\centering
\begin{tabular}{lccccccc}
\toprule
Model & Acc & Comp & N.C. & $\rho$ (\%) & CDF & Sparsity \\
\midrule
VGGT & 0.047 & 0.049 & 0.897 & 10 & 0.970 & 7.630 \\
VGGT & 0.047 & 0.049 & 0.899 & 30 & 0.970 & 11.580 \\
VGGT & 0.049 & 0.051 & 0.899 & 50 & 0.970 & 13.870 \\
VGGT & 0.048 & 0.050 & 0.898 & 40 & 0.950 & 16.550 \\
VGGT & 0.051 & 0.054 & 0.893 & 60 & 0.930 & 22.310 \\
VGGT & 0.049 & 0.051 & 0.897 & 50 & 0.900 & 25.540 \\
VGGT & 0.050 & 0.052 & 0.894 & 50 & 0.880 & 28.170 \\
\bottomrule
\end{tabular}

\caption{Full results for NRGBD~\cite{azinovic2022nrgbd}}
\end{table*}

\begin{table*}[p]
\centering
\begin{tabular}{lccccccc}
\toprule
Model & RRA@30 & RTA@30 & AUC@30 & $\rho$ (\%) & CDF & Sparsity \\
\midrule
VGGT & 0.977 & 0.763 & 0.760 & 10 & 0.970 & 7.630 \\
VGGT & 0.977 & 0.762 & 0.759 & 30 & 0.970 & 11.580 \\
VGGT & 0.976 & 0.762 & 0.759 & 50 & 0.970 & 13.870 \\
VGGT & 0.976 & 0.759 & 0.756 & 40 & 0.950 & 16.550 \\
VGGT & 0.975 & 0.754 & 0.751 & 60 & 0.930 & 22.310 \\
VGGT & 0.976 & 0.752 & 0.749 & 50 & 0.900 & 25.540 \\
VGGT & 0.975 & 0.750 & 0.746 & 50 & 0.880 & 28.170 \\
\bottomrule
\end{tabular}

\caption{Full results for Real Estate 10K~\cite{zhou2018re10k}}
\end{table*}

\begin{table*}[p]
\centering
\begin{tabular}{lccccccc}
\toprule
Model & ATE & RPEtrans & RPErot & $\rho$ (\%) & CDF & Sparsity \\
\midrule
VGGT & 0.035 & 0.015 & 0.382 & 10 & 0.970 & 7.560 \\
VGGT & 0.035 & 0.015 & 0.382 & 30 & 0.970 & 11.490 \\
VGGT & 0.036 & 0.016 & 0.393 & 50 & 0.970 & 13.690 \\
VGGT & 0.036 & 0.015 & 0.384 & 40 & 0.950 & 16.390 \\
VGGT & 0.036 & 0.016 & 0.387 & 60 & 0.930 & 22.170 \\
VGGT & 0.036 & 0.016 & 0.389 & 50 & 0.900 & 25.370 \\
VGGT & 0.036 & 0.016 & 0.392 & 50 & 0.880 & 28.010 \\
\bottomrule
\end{tabular}

\caption{Full results for ScanNet~\cite{dai2017scannet} (first 90 frames with temporal stride of 3).}
\end{table*}

\begin{table*}[p]
\centering
\begin{tabular}{lccccccc}
\toprule
Model & ATE & RPEtrans & RPErot & $\rho$ (\%) & CDF & Sparsity \\
\midrule
VGGT & 0.012 & 0.010 & 0.352 & 10 & 0.970 & 7.560 \\
VGGT & 0.012 & 0.010 & 0.352 & 30 & 0.970 & 11.490 \\
VGGT & 0.012 & 0.010 & 0.353 & 50 & 0.970 & 13.690 \\
VGGT & 0.012 & 0.010 & 0.354 & 40 & 0.950 & 16.390 \\
VGGT & 0.013 & 0.011 & 0.356 & 60 & 0.930 & 22.170 \\
VGGT & 0.013 & 0.011 & 0.358 & 50 & 0.900 & 25.370 \\
VGGT & 0.013 & 0.011 & 0.359 & 50 & 0.880 & 28.010 \\
\bottomrule
\end{tabular}

\caption{Full results for TUM~\cite{sturm2012tumdynamic}.}
\end{table*}

\begin{table*}[p]
\centering
\small
\setlength{\tabcolsep}{2pt}
\begin{tabular}{lcccccccccccccc}
\toprule
Model & RRA@5 & RRA@15 & RRA@30 & RTA@5 & RTA@15 & RTA@30 & AUC@5 & AUC@15 & AUC@30 & ATE & $\rho$ (\%) & CDF & Sparsity \\
\midrule
VGGT & 0.840 & 0.895 & 0.913 & 0.817 & 0.885 & 0.911 & 0.768 & 0.852 & 0.880 & 0.024 & 10 & 0.970 & 7.273 \\
VGGT & 0.836 & 0.894 & 0.912 & 0.815 & 0.884 & 0.910 & 0.764 & 0.851 & 0.879 & 0.025 & 30 & 0.970 & 11.093 \\
VGGT & 0.828 & 0.890 & 0.910 & 0.811 & 0.882 & 0.909 & 0.757 & 0.847 & 0.877 & 0.025 & 40 & 0.950 & 15.976 \\
VGGT & 0.741 & 0.859 & 0.894 & 0.753 & 0.860 & 0.898 & 0.660 & 0.812 & 0.858 & 0.024 & 50 & 0.700 & 44.533 \\
VGGT & 0.779 & 0.873 & 0.900 & 0.778 & 0.870 & 0.902 & 0.701 & 0.827 & 0.866 & 0.024 & 50 & 0.800 & 36.316 \\
VGGT & 0.798 & 0.880 & 0.905 & 0.791 & 0.875 & 0.905 & 0.722 & 0.835 & 0.871 & 0.024 & 50 & 0.850 & 31.045 \\
VGGT & 0.808 & 0.883 & 0.906 & 0.797 & 0.877 & 0.907 & 0.732 & 0.838 & 0.873 & 0.024 & 50 & 0.880 & 27.458 \\
\bottomrule
\end{tabular}

\caption{Full results for Tanks \& Temples~\cite{knapitsch2017tanksAndTemples} (50 frames).}
\end{table*}

\begin{table*}[p]
\centering
\small
\setlength{\tabcolsep}{2pt}
\begin{tabular}{lcccccccccccccc}
\toprule
Model & RRA@5 & RRA@15 & RRA@30 & RTA@5 & RTA@15 & RTA@30 & AUC@5 & AUC@15 & AUC@30 & ATE & $\rho$ (\%) & CDF & Sparsity \\
\midrule
VGGT & 0.837 & 0.893 & 0.912 & 0.809 & 0.879 & 0.908 & 0.761 & 0.847 & 0.877 & 0.016 & 10 & 0.970 & 7.273 \\
VGGT & 0.836 & 0.893 & 0.912 & 0.808 & 0.879 & 0.908 & 0.759 & 0.846 & 0.877 & 0.016 & 30 & 0.970 & 11.093 \\
VGGT & 0.832 & 0.891 & 0.910 & 0.806 & 0.879 & 0.908 & 0.753 & 0.845 & 0.876 & 0.017 & 40 & 0.950 & 15.976 \\
VGGT & 0.740 & 0.856 & 0.891 & 0.746 & 0.854 & 0.894 & 0.653 & 0.806 & 0.855 & 0.017 & 50 & 0.700 & 44.533 \\
VGGT & 0.780 & 0.872 & 0.899 & 0.772 & 0.865 & 0.900 & 0.696 & 0.823 & 0.864 & 0.016 & 50 & 0.800 & 36.316 \\
VGGT & 0.798 & 0.877 & 0.902 & 0.784 & 0.869 & 0.903 & 0.716 & 0.830 & 0.867 & 0.017 & 50 & 0.850 & 31.045 \\
VGGT & 0.808 & 0.882 & 0.905 & 0.790 & 0.872 & 0.905 & 0.727 & 0.834 & 0.870 & 0.017 & 50 & 0.880 & 27.458 \\
\bottomrule
\end{tabular}

\caption{Full results for Tanks \& Temples~\cite{knapitsch2017tanksAndTemples} (100 frames).}
\end{table*}

\begin{table*}[p]
\centering
\small
\setlength{\tabcolsep}{2pt}
\begin{tabular}{lcccccccccccccc}
\toprule
Model & RRA@5 & RRA@15 & RRA@30 & RTA@5 & RTA@15 & RTA@30 & AUC@5 & AUC@15 & AUC@30 & ATE & $\rho$ (\%) & CDF & Sparsity \\
\midrule
VGGT & 0.829 & 0.887 & 0.907 & 0.798 & 0.873 & 0.904 & 0.747 & 0.839 & 0.871 & 0.012 & 10 & 0.970 & 7.273 \\
VGGT & 0.826 & 0.886 & 0.906 & 0.797 & 0.873 & 0.903 & 0.745 & 0.838 & 0.870 & 0.012 & 30 & 0.970 & 11.093 \\
VGGT & 0.821 & 0.885 & 0.905 & 0.794 & 0.872 & 0.903 & 0.739 & 0.836 & 0.869 & 0.012 & 40 & 0.950 & 15.976 \\
VGGT & 0.734 & 0.854 & 0.889 & 0.737 & 0.850 & 0.891 & 0.643 & 0.801 & 0.851 & 0.012 & 50 & 0.700 & 44.533 \\
VGGT & 0.773 & 0.868 & 0.896 & 0.763 & 0.860 & 0.896 & 0.685 & 0.817 & 0.859 & 0.012 & 50 & 0.800 & 36.316 \\
VGGT & 0.791 & 0.874 & 0.899 & 0.775 & 0.865 & 0.899 & 0.704 & 0.824 & 0.863 & 0.012 & 50 & 0.850 & 31.045 \\
VGGT & 0.800 & 0.878 & 0.901 & 0.781 & 0.867 & 0.900 & 0.715 & 0.828 & 0.864 & 0.012 & 50 & 0.880 & 27.458 \\
\bottomrule
\end{tabular}

\caption{Full results for Tanks \& Temples~\cite{knapitsch2017tanksAndTemples} (200 frames).}
\end{table*}

\begin{table*}[p]
\centering
\small
\setlength{\tabcolsep}{2pt}
\begin{tabular}{lccccccc}
\toprule
Model & ATE & RPEtrans & RPErot & $\rho$ (\%) & CDF & Sparsity \\
\midrule
VGGT & 0.183 & 0.117 & 4.632 & 10 & 0.970 & 6.879 \\
VGGT & 0.184 & 0.118 & 4.659 & 30 & 0.970 & 10.103 \\
VGGT & 0.185 & 0.118 & 4.709 & 40 & 0.950 & 14.749 \\
VGGT & 0.192 & 0.116 & 4.851 & 50 & 0.700 & 44.131 \\
VGGT & 0.190 & 0.119 & 4.739 & 50 & 0.800 & 35.240 \\
VGGT & 0.191 & 0.117 & 4.801 & 50 & 0.850 & 29.774 \\
VGGT & 0.191 & 0.116 & 4.711 & 50 & 0.880 & 26.108 \\
\bottomrule
\end{tabular}

\caption{Full results for ScanNet~\cite{dai2017scannet} (100 frames evenly sampled from full sequence).}
\end{table*}

\begin{table*}[p]
\centering
\small
\setlength{\tabcolsep}{2pt}
\begin{tabular}{lccccccc}
\toprule
Model & ATE & RPEtrans & RPErot & $\rho$ (\%) & CDF & Sparsity \\
\midrule
VGGT & 0.176 & 0.069 & 2.485 & 10 & 0.970 & 6.879 \\
VGGT & 0.173 & 0.061 & 2.383 & 30 & 0.970 & 10.103 \\
VGGT & 0.173 & 0.060 & 2.386 & 40 & 0.950 & 14.749 \\
VGGT & 0.181 & 0.076 & 2.576 & 50 & 0.700 & 44.131 \\
VGGT & 0.178 & 0.069 & 2.522 & 50 & 0.800 & 35.240 \\
VGGT & 0.175 & 0.063 & 2.424 & 50 & 0.850 & 29.774 \\
VGGT & 0.175 & 0.064 & 2.417 & 50 & 0.880 & 26.108 \\
\bottomrule
\end{tabular}

\caption{Full results for ScanNet~\cite{dai2017scannet} (300 frames evenly sampled from full sequence).}
\end{table*}

\begin{table*}[p]
\centering
\small
\setlength{\tabcolsep}{2pt}
\begin{tabular}{lccccccc}
\toprule
Model & ATE & RPEtrans & RPErot & $\rho$ (\%) & CDF & Sparsity \\
\midrule
VGGT & 0.178 & 0.056 & 1.814 & 10 & 0.970 & 6.879 \\
VGGT & 0.180 & 0.060 & 1.868 & 30 & 0.970 & 10.103 \\
VGGT & 0.177 & 0.053 & 1.805 & 40 & 0.950 & 14.749 \\
VGGT & 0.186 & 0.063 & 2.274 & 50 & 0.700 & 44.131 \\
VGGT & 0.183 & 0.061 & 2.010 & 50 & 0.800 & 35.240 \\
VGGT & 0.182 & 0.061 & 1.923 & 50 & 0.850 & 29.774 \\
VGGT & 0.183 & 0.066 & 2.098 & 50 & 0.880 & 26.108 \\
\bottomrule
\end{tabular}

\caption{Full results for ScanNet~\cite{dai2017scannet} (500 frames evenly sampled from full sequence).}
\end{table*}

\begin{table*}[p]
\centering
\small
\setlength{\tabcolsep}{2pt}
\begin{tabular}{lccccccc}
\toprule
Model & ATE & RPEtrans & RPErot & $\rho$ (\%) & CDF & Sparsity \\
\midrule
VGGT & 0.204 & 0.056 & 1.587 & 10 & 0.970 & 6.879 \\
VGGT & 0.203 & 0.058 & 1.608 & 30 & 0.970 & 10.103 \\
VGGT & 0.206 & 0.061 & 1.678 & 40 & 0.950 & 14.749 \\
VGGT & 0.214 & 0.060 & 1.892 & 50 & 0.700 & 44.131 \\
VGGT & 0.213 & 0.059 & 1.770 & 50 & 0.800 & 35.240 \\
VGGT & 0.213 & 0.061 & 1.876 & 50 & 0.850 & 29.774 \\
VGGT & 0.212 & 0.062 & 1.899 & 50 & 0.880 & 26.108 \\
\bottomrule
\end{tabular}

\caption{Full results for ScanNet~\cite{dai2017scannet} (1000 frames evenly sampled from full sequence).}
\end{table*}

\end{document}